\documentclass[11pt]{article}

\usepackage[T1]{fontenc}
\usepackage[utf8]{inputenc}
\usepackage{lmodern}
\usepackage{microtype}

\usepackage[english]{babel}

\usepackage{amsmath,amssymb,mathtools,bm}

\usepackage{graphicx}
\usepackage[dvipsnames]{xcolor}
\graphicspath{{figures/}}
\usepackage{caption}
\usepackage{float}
\usepackage{subcaption}
\usepackage{dblfloatfix}

\usepackage{booktabs}
\usepackage{multirow}
\usepackage{siunitx}
\renewcommand{\arraystretch}{1.15}

\usepackage{enumitem}
\usepackage{wrapfig}
\setlist{nosep}
\usepackage{placeins}
\usepackage{titlesec}
\titlespacing*{\section}{0pt}{1.0ex plus .2ex}{0.6ex}
\titlespacing*{\subsection}{0pt}{0.8ex plus .2ex}{0.4ex}

\usepackage{listings}
\usepackage{natbib}

\usepackage{hyperref}
\hypersetup{
  colorlinks=true,
  linkcolor=MidnightBlue,
  citecolor=ForestGreen,
  urlcolor=RoyalBlue
}
\usepackage[a4paper,margin=1in]{geometry}
\usepackage{multicol}
\usepackage{abstract}

\usepackage{setspace}
\newcommand{\keywords}[1]{\par\noindent\textbf{Keywords: }#1}

\begin{document}
\thispagestyle{empty}

\noindent

\begin{center}

{\LARGE\bfseries Improving Machine Learning Performance with Synthetic Augmentation}
\vspace{0.6cm}

Charles DEZONS, Sami SELLAMI, Oscar NINOU, Axel PINCON, Mel SOHM\footnotemark

\footnotetext{\noindent
\texttt{charles.dezons@berkeley.edu} \quad|\quad
\texttt{sami\_sellami@berkeley.edu} \quad|\quad
\texttt{oscar.ninou@berkeley.edu} \quad|\quad
\texttt{axel.pincon@berkeley.edu} \quad|\quad
\texttt{mel\_sohm@berkeley.edu}}

\vspace{0.4cm}

{\normalsize University of California, Berkeley}

\vspace{0.3cm}

Feb 2026

\end{center}

\vspace{0.3cm}
\hrule height0.9pt
\vspace{0.3cm}

\begin{center}
\begin{abstract}
\noindent
Synthetic augmentation is increasingly used to mitigate data scarcity in financial machine learning, yet its statistical role remains poorly understood. We formalize synthetic augmentation as a modification of the effective training distribution and show that it induces a structural bias--variance trade-off: while additional samples may reduce estimation error, they may also shift the population objective whenever the synthetic distribution deviates from regions relevant under evaluation. To isolate informational gains from mechanical sample-size effects, we introduce a size-matched null augmentation and a finite-sample, non-parametric block permutation test that remains valid under weak temporal dependence. 
We evaluate this framework in both controlled Markov-switching environments and real financial datasets, including high-frequency option trade data and a daily equity panel. Across generators spanning bootstrap, copula-based models, variational autoencoders, diffusion models, and TimeGAN, we vary augmentation ratio, model capacity, task type, regime rarity, and signal-to-noise. We show that synthetic augmentation is beneficial only in variance-dominant regimes, such as persistent volatility forecasting-while it deteriorates performance in bias-dominant settings, including near-efficient directional prediction. Rare-regime targeting can improve domain-specific metrics but may conflict with unconditional permutation inference. Our results provide a structural perspective on when synthetic data improves financial learning performance and when it induces persistent distributional distortion.

\end{abstract}

\vspace{0.15em}
\end{center}

\vspace{0.3cm}
\hrule height0.5pt
\vspace{0.3cm}

\begin{multicols}{2}
\section{Introduction}
\label{sec:intro_lit}
\vspace{0.1cm}

Synthetic augmentation is often presented as a generic remedy for data scarcity.
In financial learning problems, however, adding synthetic observations is not a neutral operation:
it \emph{modifies the effective training distribution}.
If a learner is trained on a mixture
$P_\alpha = (1-\alpha)P_{\mathrm{real}} + \alpha P_{\mathrm{synth}}$,
then even with infinite data the population target shifts from the real-data minimizer
to the mixture minimizer.
As a result, synthetic augmentation is structurally a \emph{bias--variance trade-off}:
it can reduce estimation error by increasing effective sample size,
but it can also induce non-vanishing population shift whenever
$P_{\mathrm{synth}}$ deviates from the regions relevant under the evaluation distribution.
This tension is particularly acute in finance, where signal is weak, dependence is strong,
regimes shift, and tail events are economically decisive.
\vspace{0.1cm}

\noindent
Machine learning has become a standard toolkit in empirical finance, with applications ranging from
return forecasting and volatility prediction to portfolio construction, hedging, and risk management.
A prominent benchmark is the comparative asset-pricing analysis of \citet{gu2020empirical},
which frames risk-premium measurement as a prediction problem and documents sizable economic gains from nonlinear methods.
These gains, however, also sharpen concerns about overfitting, temporal dependence, and the credibility of out-of-sample
comparisons in settings with weak signal and structural change.
Unlike domains such as computer vision or natural language processing, financial datasets are characterized by
limited effective sample size, heavy-tailed distributions, rare but critical events (crashes, liquidity shocks),
and a dynamic data-generating process as market participants adapt and macro regimes evolve.
Under these conditions, the practical question is not whether synthetic data \emph{looks realistic},
but whether it contributes \emph{incremental predictive information} in the regime where performance matters.
\vspace{0.1cm}

\noindent
A key difficulty is evaluation.
If a synthetic-augmented model is compared only to a real-only baseline,
any performance difference conflates two channels:
(i) a mechanical sample-size effect (variance reduction) and
(ii) an informational effect (the generator adding structure aligned with $P_{\mathrm{test}}$).
This paper studies synthetic augmentation through a controlled counterfactual:
we compare augmentation by $\tilde{\mathcal D}_n \sim P_{\mathrm{synth}}$
to a \emph{size-matched null augmentation} $\tilde{\mathcal D}_n^{\mathrm{null}}$
that preserves low-level properties (marginals, scale, and where relevant temporal structure),
but destroys predictive alignment.
This isolates whether synthetic data adds usable information beyond what any
non-informative increase in sample size would deliver.
\vspace{0.1cm}

\noindent
Our objective is therefore structural rather than benchmark-specific.
Rather than asking whether synthetic data ``works'' in a particular dataset,
we aim to understand:
(i) when augmentation is variance-dominant versus bias-dominant;
(ii) how usefulness depends on task type (near-efficient directional prediction versus persistent volatility);
(iii) how effectiveness varies with regime rarity and tail behavior; and
(iv) how generator class (support-preserving resampling, parametric dependence models,
deep generative models) interacts with model capacity and temporal dependence constraints.
To make these questions testable, we formalize the augmentation effect as an out-of-sample risk differential
and introduce a finite-sample, non-parametric (block) permutation test valid under weak dependence,
as developed in Section~\ref{sec:framework}.
\vspace{0.1cm}

\noindent
A natural baseline for synthetic augmentation is resampling.
The bootstrap of \citet{efron1979bootstrap} motivates generating pseudo-samples from the empirical distribution
to approximate sampling uncertainty.
Because financial observations are dependent, dependence-aware resampling is essential:
\citet{kunsch1989jackknife} extend jackknife/bootstrap logic to stationary sequences using blocks,
and \citet{politis1994stationary} propose the stationary bootstrap with random block lengths to better preserve stationarity.
In parallel, the synthetic oversampling literature in machine learning emphasizes targeting rare states or classes:
SMOTE (\citet{chawla2002smote}) generates synthetic minority examples and provides a canonical illustration that
augmentation can help when loss is sensitive to underrepresented outcomes.
\vspace{0.1cm}

\noindent
Modern augmentation frequently relies on learned generators.
TimeGAN (\citet{yoon2019timegan}) is a landmark time-series GAN that combines adversarial training with a supervised component
to better preserve temporal dynamics.
In finance, Quant GANs (\citet{wiese2020quantgan}) propose architectures designed to reproduce stylized facts and dependence properties of market data.
Dogariu et al.\ (\citet{dogariu2022realistic}) emphasize the challenge of preserving cross-sectional co-movement
and advocate evaluation using both statistical and task-level criteria.
Recent diffusion-based approaches broaden this toolkit:
TimeGrad (\citet{rasul2021timegrad}) introduces an autoregressive diffusion formulation for multivariate probabilistic forecasting,
and \citet{takahashi2024diffusionfinance} propose diffusion-based generation for financial time series using wavelet/image representations.
Bridging deep generation and resampling, \citet{dahl2022ganbootstrap} propose a GAN-based bootstrap for dependent processes
and evaluate finite-sample performance relative to block-bootstrap baselines, including an application to the Sharpe ratio.
\vspace{0.1cm}

\noindent
Beyond raw predictive accuracy, financial ML often faces interpretability, governance, and bias constraints.
In consumer credit, \citet{bell2025fairness} develop an interpretable approach to mortgage lending that embeds fairness constraints
and produces decision-level feature attributions suitable for adverse-action explanations.
In asset pricing, \citet{bell2024glass} show that interpretable ``glass box'' models can deliver strong performance in corporate bond return prediction
while retaining economic transparency.
Recent work also demonstrates how generative AI and text pipelines create new measurement systems in finance and macroeconomics:
\citet{kakhbod2024measuring} develop a text-based measure of innovation displacement,
\citet{fedyk2024ai} study perception biases in AI-generated investment advice,
\citet{kakhbod2025fed} extract the Federal Reserve's inflation attribution using LLMs,
and \citet{hochberg2023patents} use causal text estimation to study gender gaps in patent under-citation.
These studies motivate careful validation of any synthetic or AI-generated inputs, as biases or fidelity failures can propagate into economic inference.
\vspace{0.1cm}

\noindent
Our contribution is to provide a statistically rigorous evaluation of synthetic augmentation in financial learning problems,
formalizing augmentation as a change in the effective training distribution and isolating informational gains from mechanical sample-size effects
via a size-matched null augmentation and a finite-sample, nonparametric testing framework.
We evaluate this framework in both controlled toy environments and real financial datasets, and we document how outcomes depend jointly on
model variance, regime under-representation, temporal dependence, and the fidelity of the generative mechanism.

\section{Framework}\label{sec:framework}

\subsection*{Object of the Test}

Let $A(\cdot)$ denote a fixed learning algorithm mapping a dataset into a predictor.
All architectural, optimization and hyperparameter choices are held fixed.
\vspace{0.1cm}

\noindent
Given a real dataset $\mathcal D_m \sim P_{\mathrm{real}}$
and a synthetic augmentation $\tilde{\mathcal D}_n \sim P_{\mathrm{synth}}$,
define the augmented estimator

\begin{equation}
\hat f^{\mathrm{syn}}_n = A(\mathcal D_m \cup \tilde{\mathcal D}_n).
\label{eq:augmented-estimator}
\end{equation}

\noindent
To isolate the informational contribution of synthetic data, we construct
a null augmentation $\tilde{\mathcal D}_n^{\mathrm{null}}$
of identical size $n$ such that it preserves low-level properties
(marginal distributions, scale, possibly temporal structure),
but contains no additional predictive signal about $Y$.
\vspace{0.1cm}

\noindent
Examples include:
\vspace{0.1cm}

\begin{itemize}
\item label permutation,
\item temporal shuffling destroying dependence structure,
\item blockwise resampling with destroyed cross-sectional alignment.
\end{itemize}
\vspace{0.1cm}

\noindent
We define

\begin{equation}
\hat f^{\mathrm{null}} = A(\mathcal D_m \cup \tilde{\mathcal D}_n^{\mathrm{null}}).
\label{eq:null-estimator}
\end{equation}

\noindent
Both estimators are trained under identical conditions.
The only difference is whether the added sample carries structured signal.
Let $\mathcal D_{\mathrm{test}} = \{(x_t,y_t)\}_{t=1}^T$
be an out-of-sample dataset drawn from $P_{\mathrm{test}}$.
For a loss $\ell$, define the loss differential
\vspace{0.1cm}

\noindent
\begin{equation}
d_t
=
\ell(\hat f^{\mathrm{null}}(x_t), y_t)
-
\ell(\hat f^{\mathrm{syn}}_n(x_t), y_t).
\label{eq:loss-differential}
\end{equation}

\noindent
The observed performance differential is

\begin{equation}
\hat\delta = \frac{1}{T}\sum_{t=1}^T d_t.
\label{eq:observed-differential}
\end{equation}

\noindent
The parameter of interest is

\begin{equation}
\delta = \mathbb E_{P_{\mathrm{test}}}[d_t].
\label{eq:parameter-delta}
\end{equation}

\subsection*{Interpretation of the Null Hypothesis}
\vspace{0.1cm}

We define the risk for an estimator $f$ as,
\begin{equation}
R(f) = \mathbb{E}_{P_{test}}[l(f(X), Y)]
\label{eq:Risk}
\end{equation}
\noindent
We test

\begin{equation}
H_0 : \delta = 0
\quad \text{vs} \quad
H_1 : \delta > 0.
\label{eq:null-hypothesis}
\end{equation}

\noindent
Importantly, this is not a test of equality between
$P_{\mathrm{real}}$ and $P_{\mathrm{synth}}$.
It is a test of incremental predictive information
conditional on the learning algorithm and evaluation distribution.
Under $H_0$, synthetic augmentation does not alter
the expected out-of-sample loss relative to a size-matched null augmentation.
Any observed difference is attributable to random fluctuations
in the finite test sample.
Thus, the null hypothesis formalizes:
\begin{equation}
\mathbb E\!\left[
R_{P_{\mathrm{test}}}(\hat f^{\mathrm{syn}}_n)
-
R_{P_{\mathrm{test}}}(\hat f^{\mathrm{null}})
\right] = 0.
\label{eq:null-risk-equivalence}
\end{equation}

\noindent
The conditioning on $\mathcal D_m$ is implicit:
the test assesses marginal informational contribution
given the observed real dataset.
\vspace{0.1cm}

\subsection*{Bias-Variance Decomposition of the Effect}
\vspace{0.1cm}

To understand what $\delta$ captures at a structural level,
we introduce the population risks under the real and mixed training distributions.
Let
\begin{equation}
P_\alpha = (1-\alpha)P_{\mathrm{real}} + \alpha P_{\mathrm{synth}},
\qquad
\alpha = \frac{n}{m+n}.
\label{eq:mixture-distribution}
\end{equation}

\noindent
Denote the corresponding population minimizers
\begin{align}
f_{\mathrm{real}}^*
&=
\arg\min_{f \in \mathcal F} R_{P_{\mathrm{real}}}(f),
\label{eq:real-population-minimizer}
\\
f_\alpha^*
&=
\arg\min_{f \in \mathcal F} R_{P_\alpha}(f).
\label{eq:mixture-population-minimizer}
\end{align}

\noindent
As $m,n \to \infty$ with fixed $\alpha$,
\begin{align}
\hat f^{\mathrm{syn}}_n &\to f_\alpha^*,
\label{eq:syn-convergence}
\\
\hat f_m &\to f_{\mathrm{real}}^*.
\label{eq:real-convergence}
\end{align}

\noindent
Synthetic augmentation therefore changes
the population objective itself.
The target of estimation shifts from $f_{\mathrm{real}}^*$ to $f_\alpha^*$.
We now decompose the expected risk differential
\begin{equation}
\Delta R(m,n)
=
\mathbb E\!\left[
R_{P_{\mathrm{test}}}(\hat f^{\mathrm{syn}}_n)
-
R_{P_{\mathrm{test}}}(\hat f_m)
\right].
\label{eq:delta-risk}
\end{equation}

\noindent
When the null augmentation leaves the real sample unchanged in expectation,
$\hat f^{\mathrm{null}}$ coincides with $\hat f_m$,
so that $\delta$ estimates $\Delta R(m,n)$.
Add and subtract the corresponding population optima:
\begin{equation}
\begin{aligned}
\Delta R
&=
\big[
R_{P_{\mathrm{test}}}(f_\alpha^*)
-
R_{P_{\mathrm{test}}}(f_{\mathrm{real}}^*)
\big]
\\
&\quad
+
\mathbb E\big[
R_{P_{\mathrm{test}}}(\hat f^{\mathrm{syn}}_n)
-
R_{P_{\mathrm{test}}}(f_\alpha^*)
\big]
\\
&\quad
-
\mathbb E \big[
R_{P_{\mathrm{test}}}(\hat f_m)
-
R_{P_{\mathrm{test}}}(f_{\mathrm{real}}^*)
\big].
\end{aligned}
\label{eq:bias-variance-decomposition}
\end{equation}

\noindent
The first term is a population shift.
It persists asymptotically whenever $\alpha > 0$
and vanishes only if $P_{\mathrm{synth}} = P_{\mathrm{real}}$
on regions relevant under $P_{\mathrm{test}}$.
The second term (the two expectations) captures estimation error.
Under standard regularity conditions,
its magnitude decreases as the effective sample size increases.
\vspace{0.1cm}

\noindent
Synthetic data is beneficial if and only if
variance reduction exceeds population shift.
This inequality formalizes the central trade-off:
augmentation reduces stochastic error
but may introduce systematic misspecification.\citep{dao2019kernel}
\vspace{0.1cm}

\subsection*{Exchangeability and Finite-Sample Validity}
\vspace{0.1cm}

\noindent
Under $H_0$, the joint distribution of
$(\ell(\hat f^{\mathrm{syn}}_n(x_t),y_t),
  \ell(\hat f^{\mathrm{null}}(x_t),y_t))$
is invariant to permutation of the two predictors,
which implies sign-exchangeability of $\{d_t\}$.
\begin{equation}
(d_1,\dots,d_T)
\stackrel{d}{=}
(s_1 d_1,\dots,s_T d_T),
\label{eq:sign-exchangeability}
\end{equation}

\noindent
where $s_t \in \{-1,+1\}$ are i.i.d.
Exchangeability is the minimal assumption required
for exact finite-sample validity of a permutation test.
No Gaussianity, independence, or asymptotic approximation is required.
The permutation statistic is constructed as
\begin{equation}
\hat\delta^{(b)}
=
\frac{1}{T}\sum_{t=1}^T s_t^{(b)} d_t.
\label{eq:permutation-statistic}
\end{equation}

\noindent
The one-sided p-value is

\begin{equation}
p
=
\frac{1 + \#\{b : \hat\delta^{(b)} \ge \hat\delta\}}
{B+1}.
\label{eq:p-value}
\end{equation}
\vspace{0.1cm}

\noindent
Conditional on $\{d_t\}$,
this test controls type-I error exactly.
\vspace{0.1cm}

\subsection*{Time-Series Dependence and Block Permutation}
\vspace{0.1cm}

In financial applications,
$\{d_t\}$ is typically serially dependent.
Pointwise sign flipping breaks temporal dependence
and invalidates exchangeability.
To preserve weak dependence,
we partition the test set into $K$ non-overlapping blocks
of size $b$:
\begin{equation}
B_k = \{ t : (k-1)b < t \le kb \}.
\label{eq:block-definition}
\end{equation}

\noindent
Define blockwise aggregates
\begin{equation}
D_k = \sum_{t \in B_k} d_t.
\label{eq:block-aggregate}
\end{equation}

\noindent
Permutation is then performed at the block level:
\begin{equation}
\hat\delta^{(b)}
=
\frac{1}{T}
\sum_{k=1}^K s_k^{(b)} D_k.
\label{eq:block-permutation-statistic}
\end{equation}

\noindent
Under a standard mixing condition,
where dependence between observations decays with lag,
\begin{equation}
\alpha(k) \to 0 \quad \text{as } k \to \infty,
\label{eq:mixing-condition}
\end{equation}

\noindent
block permutation yields asymptotically valid inference
provided $b \to \infty$ and $b/T \to 0$.
This construction preserves intra-block autocorrelation
while allowing inter-block sign exchangeability.
The procedure can be interpreted as a conditional randomization test,
where the null hypothesis states that the structural component
of the synthetic augmentation carries no additional predictive signal.
\vspace{0.1cm}

\subsection*{What the Test Identifies}
\vspace{0.1cm}

The test identifies whether the synthetic distribution
modifies the learned predictor in a way
that improves expected performance under $P_{\mathrm{test}}$.
It does \emph{not} test:
\vspace{0.1cm}

\begin{itemize}
\item whether $P_{\mathrm{synth}} = P_{\mathrm{real}}$,
\item whether the generator is realistic in a generative sense,
\item whether synthetic data improves training loss.
\end{itemize}
\vspace{0.1cm}

\noindent
It isolates the causal effect of replacing
a non-informative augmentation
by a structured synthetic augmentation,
holding sample size fixed.
This distinction is essential.
An augmentation may appear realistic,
yet fail to improve prediction
if its structure is misaligned with $P_{\mathrm{test}}$.
\vspace{0.1cm}

\subsection*{When Does Augmentation Fail?}
\vspace{0.1cm}

The failure mechanism is driven by a mismatch between the
training mixture $P_\alpha$ and the evaluation distribution
$P_{\mathrm{test}}$.
Suppose there exists a measurable region $\mathcal A \subset \mathcal Z$
such that
\begin{equation}
P_{\mathrm{test}}(\mathcal A) > 0,
\qquad
P_{\mathrm{synth}}(\mathcal A) \approx 0.
\label{eq:region-mismatch}
\end{equation}

\noindent
Then the mixed training distribution satisfies
\begin{equation}
P_\alpha(\mathcal A)
=
(1-\alpha)P_{\mathrm{real}}(\mathcal A),
\label{eq:mixture-mass-region}
\end{equation}

\noindent
so that relative to $P_{\mathrm{real}}$,
the mass assigned to $\mathcal A$ is reduced
by a factor $(1-\alpha)$.
The population optimizer under the mixture,
\begin{equation}
f_\alpha^*
=
\arg\min_{f \in \mathcal F}
\mathbb E_{P_\alpha}[\ell(f,Z)],
\label{eq:mixture-optimizer-z}
\end{equation}

\noindent
therefore places less emphasis on fitting $\mathcal A$
than $f_{\mathrm{real}}^*$ does.
If the loss is large or highly variable on $\mathcal A$,
and if $\mathcal A$ carries non-negligible weight under
$P_{\mathrm{test}}$,
then the risk under $P_{\mathrm{test}}$ increases:
\begin{equation}
R_{P_{\mathrm{test}}}(f_\alpha^*)
>
R_{P_{\mathrm{test}}}(f_{\mathrm{real}}^*).
\label{eq:risk-increase}
\end{equation}

\noindent
Consequently,
\begin{equation}
\begin{aligned}
\lim_{m,n \to \infty \text{ with } \alpha \text{ fixed}} \Delta R(m,n)
&=
R_{P_{\mathrm{test}}}(f_\alpha^*)
-
R_{P_{\mathrm{test}}}(f_{\mathrm{real}}^*)
\\
&> 0.
\end{aligned}
\label{eq:asymptotic-positive-risk}
\end{equation}

\noindent
In this regime,
the asymptotic bias induced by distributional distortion
dominates any finite-sample variance reduction.
\vspace{0.1cm}

\subsection*{Rare Regimes}
\vspace{0.1cm}

Let $\mathcal A \subset \mathcal Z$
be a measurable region corresponding to
a rare or stressed regime. 
Define the conditional risk
\begin{equation}
R_{P_{\mathrm{test}}|\mathcal A}(f)
=
\mathbb E[\ell(f,Z) \mid Z \in \mathcal A].
\label{eq:conditional-risk}
\end{equation}

\noindent
The same permutation framework applies
to the restricted loss sequence
$\{d_t \mathbf 1_{Z_t \in \mathcal A}\}$,
or to a rarity-weighted loss.
This allows the test to detect
whether synthetic data contributes
information specifically in low-probability regions,
where variance is typically largest.
\vspace{0.1cm}

\subsection*{Dependence on the Learning Algorithm}
\vspace{0.1cm}

All results are conditional on $A$ and $\mathcal F$.
Synthetic information is algorithm-relative.
Two hypothesis classes may respond differently
to the same augmentation.
A high-capacity model may detect subtle mismatches
between $P_{\mathrm{synth}}$ and $P_{\mathrm{real}}$,
while a constrained model may average them out.
Therefore, any empirical claim must be indexed by
\begin{equation}
(P_{\mathrm{synth}}, \mathcal F, P_{\mathrm{test}}).
\label{eq:algorithm-relative-triple}
\end{equation}

\noindent
There is no algorithm-free notion of synthetic usefulness.

\section{Data Description}\label{sec:data}
\vspace{0.1cm}

Our empirical analysis relies on two complementary datasets spanning two distinct market microstructures and prediction regimes. The first is a \emph{tick-by-tick options trade tape} for SPY options (February 2021 to February 2026), which provides ultra-high-frequency executions with rich microstructure features and a short-horizon directional classification target. The second is a \emph{daily equity panel} sourced from Databento (January 2020 to January 2024) for five large-cap U.S.\ equities, which supports both directional and volatility-related prediction tasks at daily frequency. Their sharp contrast in granularity, dependence structure, and signal-to-noise makes them well-suited to studying when synthetic augmentation helps---and when it fails.
\vspace{0.1cm}

\subsection*{Tick-by-Tick Options Trade Tape}\label{sec:data:tick}
\vspace{0.1cm}

The first dataset is a \emph{tick-by-tick trade tape} recording every executed transaction on SPY options contracts from February 2021 to February 2026. The raw file contains approximately \textbf{160.3\,M} rows across 498 distinct option contracts and 14 fields, including nanosecond-precision timestamps (\texttt{ts\_recv}, \texttt{ts\_event}), transaction price, trade size, and aggressor side (buyer- or seller-initiated). After restricting to regular-session hours (9:30--16:00\,ET), removing trades with unknown aggressor side, and winsorising extreme trade sizes at the 99.9th percentile (1\,200 shares), the cleaned dataset retains \textbf{141.5\,M} ticks.
\vspace{0.1cm}

\noindent
The prediction target is a \emph{classification} problem: predict the direction of the 3-tick forward log-return,
$y_t = \text{sign}(r_{t+1:t+3}) \in \{-1,\, 0,\, +1\}$.
This task is dominated by microstructure noise and rare-event dependence: most ticks correspond to negligible price moves, while the economically meaningful observations (large trades, volatility bursts, regime transitions) are sparse.

\begin{figure*}[!t]
\centering
\includegraphics[width=\linewidth]{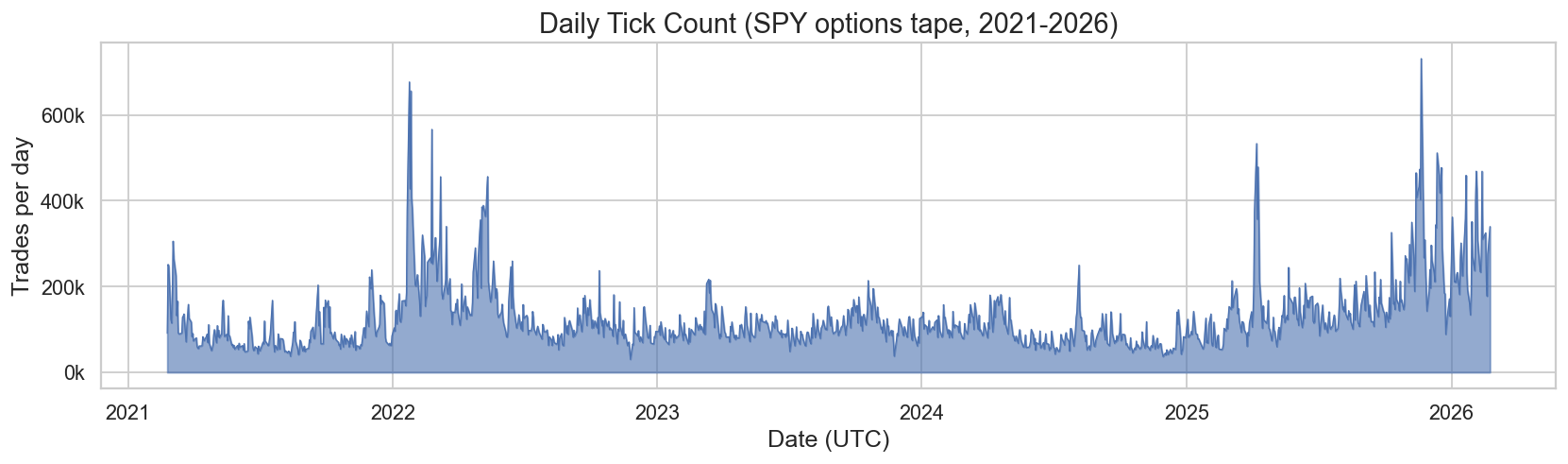}
\caption{Daily tick count for SPY options (2021--2026). Spikes correspond to major volatility events.}\label{fig:tick_daily}
\end{figure*}

\paragraph{Coverage and activity.}
Figure~\ref{fig:tick_daily} shows the daily tick count over the five-year sample. Activity averages around 128\,K trades per day (median 105\,K) but exhibits dramatic spikes, up to 732\,K trades per day-during episodes of market stress.

\begin{figure*}[!t]
\centering
\begin{subfigure}[t]{0.49\linewidth}
    \centering
    \includegraphics[width=\linewidth]{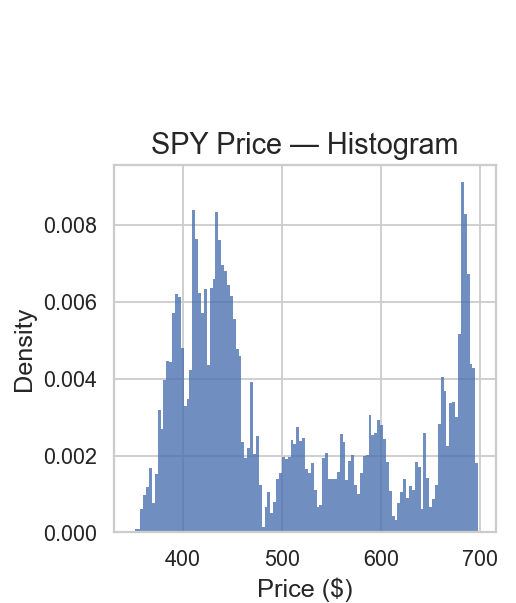}
    \caption{Histogram}
\end{subfigure}\hfill
\begin{subfigure}[t]{0.49\linewidth}
    \centering
    \includegraphics[width=\linewidth]{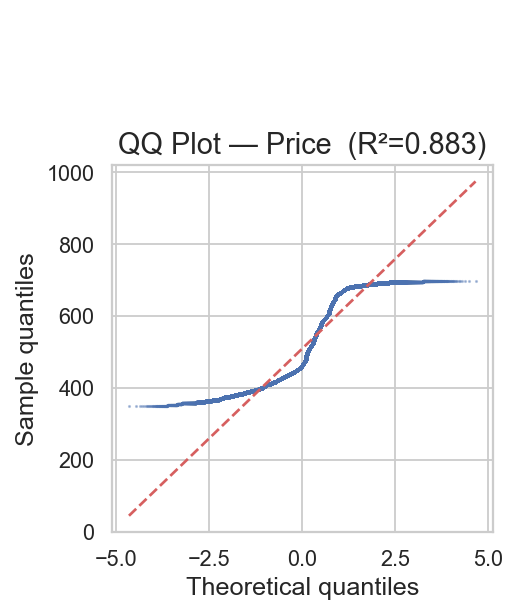}
    \caption{Q-Q plot ($R^2\!=\!0.88$)}
\end{subfigure}
\caption{SPY price distribution analysis: multi-modal structure reflecting distinct price regimes across the 2021-2026 sample.}
\label{fig:tick_price}
\end{figure*}

\paragraph{Price distribution.}
The SPY price distribution (Figure~\ref{fig:tick_price}) is \emph{multi-modal}, reflecting drift across distinct price regimes over five years (skewness $= 0.49$, excess kurtosis $= -1.25$). The KDE reveals several modes corresponding to different market epochs.

\begin{figure*}[!t]
\centering
\begin{subfigure}[t]{0.48\linewidth}
    \centering
    \includegraphics[width=\linewidth]{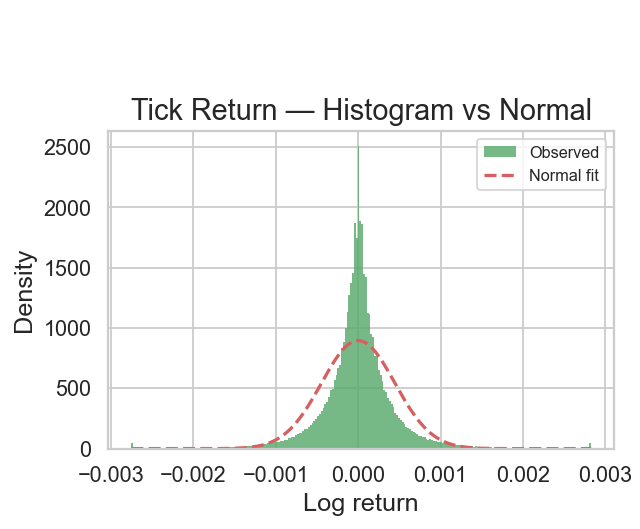}
    \caption{Histogram vs.\ Normal}
\end{subfigure}\hfill
\begin{subfigure}[t]{0.48\linewidth}
    \centering
    \includegraphics[width=\linewidth]{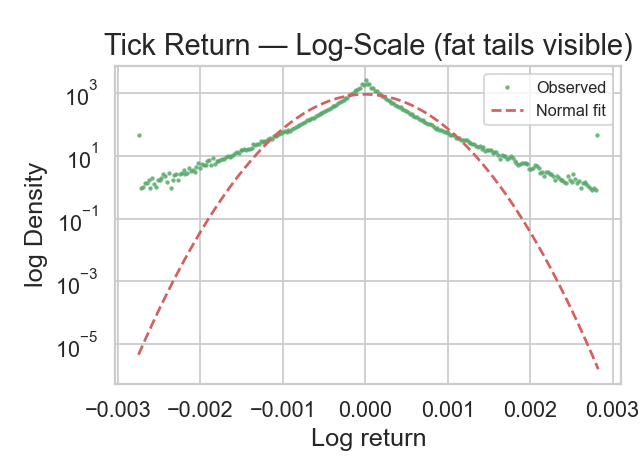}
    \caption{Log-scale density}
\end{subfigure}

\vspace{0.5cm}

\begin{subfigure}[t]{0.60\linewidth}
    \centering
    \includegraphics[width=\linewidth]{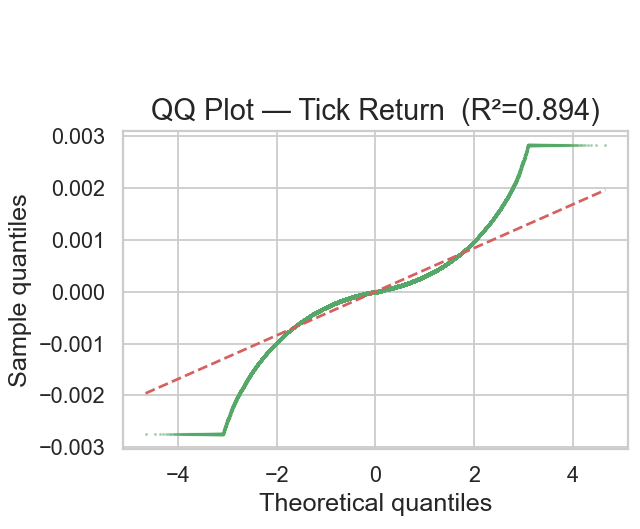}
    \caption{Q-Q plot ($R^2\!=\!0.89$)}
\end{subfigure}
\caption{Tick-to-tick log-return distribution. The log-scale view reveals heavy tails orders of magnitude above the Gaussian reference.}
\label{fig:tick_returns}
\end{figure*}

\paragraph{Tick return distribution.}
Tick-to-tick log-returns are sharply peaked around zero with extremely heavy tails: excess kurtosis exceeds 10\,000 and skewness is strongly negative ($-70.4$). Figure~\ref{fig:tick_returns} displays the return distribution on both linear and log scales alongside a Q-Q plot; the deviation from normality is orders of magnitude more severe than in daily data.

\begin{figure*}[!t]
\centering
\begin{subfigure}[t]{0.48\linewidth}
    \includegraphics[width=\linewidth]{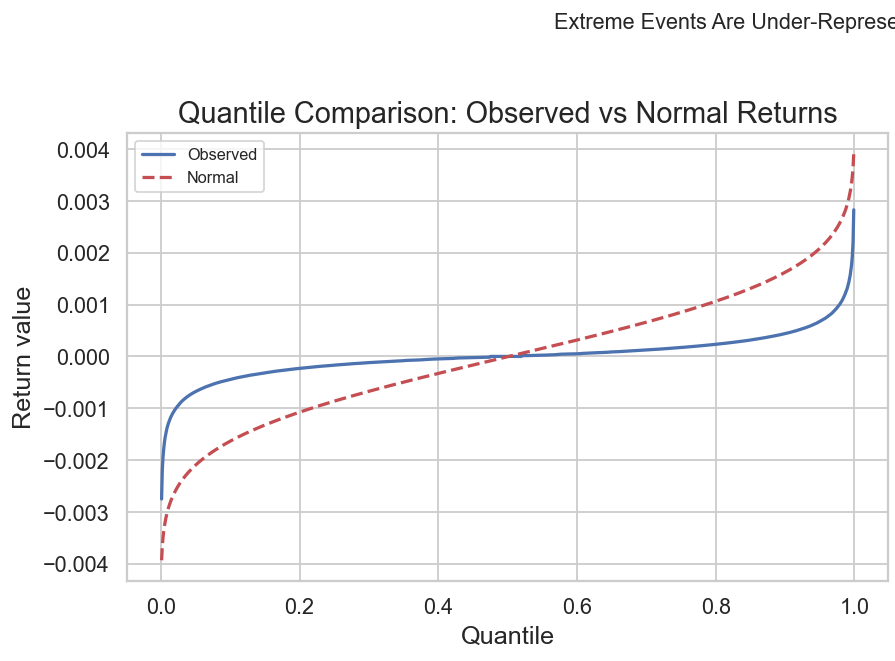}
    \caption{Full quantile function}
\end{subfigure}\hfill
\begin{subfigure}[t]{0.48\linewidth}
    \includegraphics[width=\linewidth]{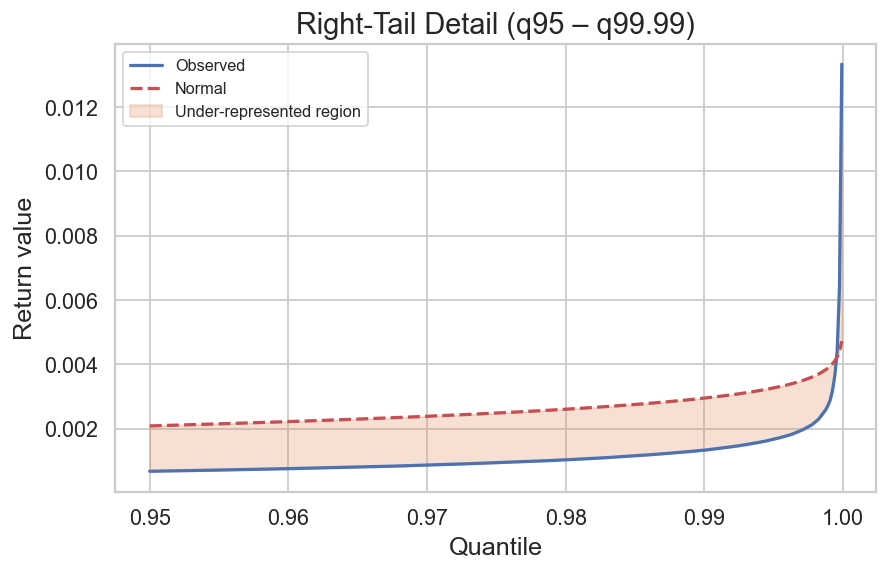}
    \caption{Right-tail zoom (q95--q99.99)}
\end{subfigure}
\caption{Quantile comparison of observed tick returns vs.\ a Normal with matched moments. The shaded area highlights where Gaussian generation under-represents extreme price moves.}\label{fig:tick_tail}
\end{figure*}

\paragraph{Tail under-representation under Gaussian calibration.}
Figure~\ref{fig:tick_tail} overlays the empirical quantile function of tick returns against that of a Normal distribution fitted to the same mean and variance. The two curves are close in the central body but diverge sharply in both tails: observed extreme returns are far larger in absolute value than Gaussian calibration would generate. This gap is precisely where augmentation quality matters: a synthetic mechanism that matches only first and second moments will systematically under-sample the events that drive risk and P\&L.

\paragraph{Trade size and microstructure features.}
Trade sizes are heavily right-skewed (skewness $= 4.86$, excess kurtosis $= 36.0$). Table~\ref{tab:tick_stats} reports distributional shape statistics for key features: order-flow imbalance (OFI), inter-trade durations, and latency measures. These variables exhibit strong non-Gaussianity and burstiness, consistent with clustered execution and intraday seasonality.

\end{multicols}
\begin{table*}[ht]
\centering
\caption{Tick dataset - Distributional shape statistics for key variables.}\label{tab:tick_stats}
\small
\begin{tabular}{@{}lrrrr@{}}
\toprule
Variable & Mean & Std & Skew & Ex.\ Kurt. \\
\midrule
\texttt{price}             & 509.6 & 106.3 & 0.49 & $-1.25$ \\
\texttt{size}              & 84.7 & 101.8 & 4.86 & 36.02 \\
\texttt{tick\_return}      & $\approx 0$ & 0.0013 & $-70.4$ & 10\,033 \\
\texttt{future\_ret\_3}    & $\approx 0$ & 0.0022 & $-40.8$ & 3\,355 \\
\texttt{inter\_trade\_dur} & $1.8 \times 10^8$ & $2.6 \times 10^9$ & 23.9 & 571 \\
\texttt{latency\_$\mu$s}   & 112.8 & 80.4 & 0.48 & 4.69 \\
\texttt{ofi\_10}           & 17.1 & 429.9 & 0.03 & 3.15 \\
\bottomrule
\end{tabular}
\end{table*}
\begin{multicols}{2}

\begin{figure*}[!t]
\centering
\begin{subfigure}[t]{0.48\linewidth}
    \includegraphics[width=\linewidth]{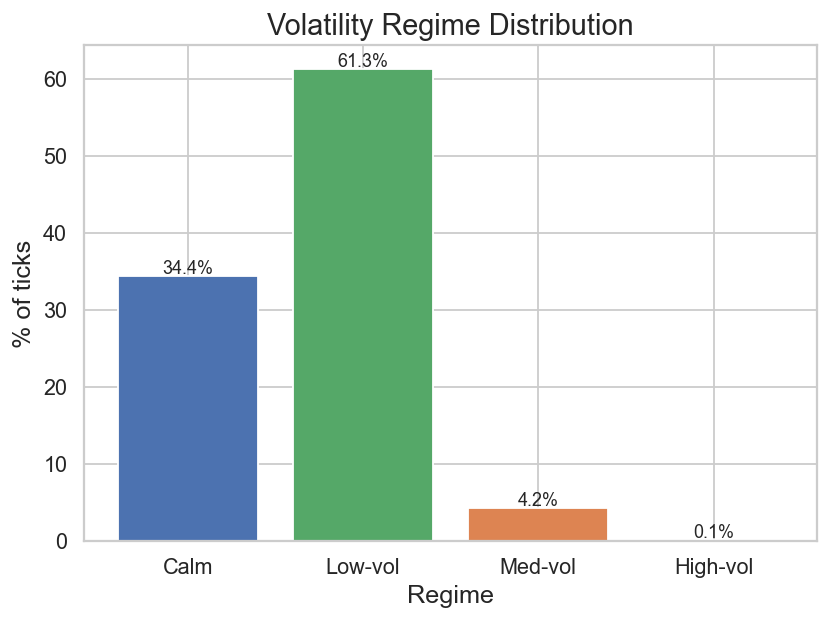}
    \caption{Regime distribution}
\end{subfigure}\hfill
\begin{subfigure}[t]{0.48\linewidth}
    \includegraphics[width=\linewidth]{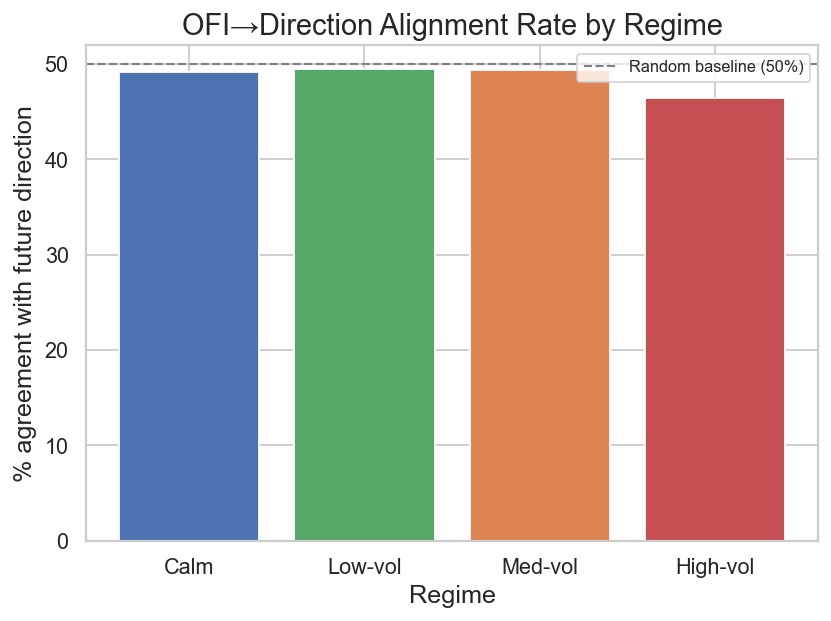}
    \caption{OFI alignment by regime}
\end{subfigure}
\caption{Left: volatility regime distribution (high-volatility ticks are extremely rare). Right: OFI-to-direction alignment rate by regime; predictive signal degrades in stressed markets.}\label{fig:tick_regimes}
\end{figure*}

\paragraph{Volatility regimes and class balance.}
Figure~\ref{fig:tick_regimes} shows a stark regime imbalance: the vast majority of ticks occur in calm/low-vol conditions, while truly stressed ticks are extremely rare. This creates a \emph{rare-regime learning} problem even at massive sample size: the challenge is not the number of observations, but the scarcity of informative regimes and the fragility of signal under stress.

\vspace{0.1cm}

\subsection*{Daily Equity Panel Dataset}\label{sec:data:equity}
\vspace{0.1cm}

The second dataset is a \emph{daily equity panel} sourced from Databento (XNAS.ITCH) for five large-cap U.S.\ equities: AAPL, MSFT, GOOGL, IBM and NVDA, covering January 2020 through January 2024 ($\approx$4{,}920 stock-day observations; 984 trading days per ticker). Fifteen features are derived from raw OHLCV data and z-score standardised within each training window to prevent look-ahead bias (Table~\ref{tab:equity_features}).
\vspace{0.1cm}

\noindent
Three tasks are defined on the same feature set:
(i) binary directional classification ($Y_t = \mathbf{1}\{r_{t+1}>0\}$, near-balanced),
(ii) one-step-ahead volatility forecasting (\texttt{vol\_5d}$_{t+1}$),
and (iii) binary detection of volatility spikes (\texttt{vol\_5d}$_{t+1} > \mathrm{p95}$, a stable rare class across folds).
This panel therefore isolates a classic daily-frequency setting where the primary difficulty is limited cross-sectional sample size and mild-to-moderate predictability (depending on the target), rather than microstructure dependence.

\end{multicols}
\begin{table*}[ht]
\centering
\caption{Daily Equity Panel: feature set.}\label{tab:equity_features}
\small
\begin{tabular}{@{}ll@{}}
\toprule
Feature & Description \\
\midrule
\texttt{ret\_lag\_k} ($k=1,2,3,5,10$) & Log-return lag $k$ \\
\texttt{vol\_kd} ($k=5,10,20$)         & Realised vol, $k$-day window \\
\texttt{mom\_kd} ($k=5,20$)            & Cumulative return, $k$ days \\
\texttt{vol\_norm\_20d}                & vol\_5d / vol\_20d \\
\texttt{high\_low\_range}              & $(H_t - L_t)/\bar H$ \\
\texttt{close\_position}               & $(C_t - L_t)/(H_t - L_t)$ \\
\texttt{rsi\_14}                       & RSI, 14-day window \\
\texttt{macd\_signal}                  & MACD signal line \\
\bottomrule
\end{tabular}
\end{table*}

\begin{table*}[ht]
\centering
\caption{Summary comparison of the two datasets.}\label{tab:comparison_2}
\small
\begin{tabular}{@{}llll@{}}
\toprule
Property & Tick Tape & Daily Equity Panel \\
\midrule
Rows (clean)  & $\sim$141.5\,M & $\sim$4{,}920 \\
Date range    & 2021--2026     & 2020--2024 \\
Frequency     & Tick-by-tick   & Daily \\
Instrument    & SPY options    & 5 equities \\
Features      & 8 numeric      & 15 numeric \\
Task type     & Classification & Clf / Reg / Rare \\
Target        & $\text{sign}(r_{t+1:t+3})$ & $\mathbf{1}\{r_{t+1}>0\}$ / vol\_5d$_{t+1}$ / $\mathbf{1}\{\text{vol\_5d}_{t+1}>\mathrm{p95}\}$ \\
Dependence    & Strong temporal microstructure & Moderate time-series dependence \\
\bottomrule
\end{tabular}
\end{table*}
\begin{multicols}{2}

\noindent
Table~\ref{tab:comparison_2} highlights the key structural contrast. The tick tape is ``large-$n$'' in row count yet dominated by rare regimes and strong sequential dependence; the daily panel is ``small-$n$'' with comparatively stable dynamics but limited cross-sectional breadth. This distinction is central for interpreting augmentation results: in the tick setting, gains (if any) must come from better coverage of \emph{rare, informative sequences}; in the daily panel, gains are more plausibly driven by variance reduction and improved learning of smooth daily relationships (especially for volatility-related targets).

\subsection*{Implications for Synthetic Augmentation}\label{sec:data:synth}
\vspace{0.1cm}

The distributional properties above imply two qualitatively different augmentation problems.
\vspace{0.1cm}

\noindent
\textbf{Tick-by-tick tape (rare regimes + dependence constraints).}
Despite the massive sample size, the tape is dominated by near-zero returns and calm-market conditions, while stressed regimes and large-move sequences are sparse (Figure~\ref{fig:tick_regimes}). Moreover, the most salient ``signal'' variables (trade sign autocorrelation, volatility clustering, intraday seasonality, order fragmentation) are inherently sequential. As a result, i.i.d.\ augmentation that matches only marginals (or even static cross-moments) is structurally mis-specified: it will destroy dependence and under-represent extremes (Figure~\ref{fig:tick_tail}). Effective augmentation in this setting must be \emph{sequence-aware} and \emph{regime-conditioned}, explicitly targeting the under-covered stressed segments while preserving joint dynamics of price, size, and order flow.
\vspace{0.1cm}

\noindent
\textbf{Daily equity panel (small sample + task-dependent signal).}
The daily panel is small in row count and contains engineered features designed to be robust out-of-sample (Table~\ref{tab:equity_features}). Here, augmentation is primarily a variance-reduction lever: adding plausible synthetic days can stabilize estimation, especially for the rare-event volatility-spike task where positive examples are scarce by construction. However, because daily returns are close to unpredictable, naive augmentation can easily become bias-dominant for directional classification if the synthetic generator injects spurious predictability. In contrast, volatility targets are typically more persistent, so carefully calibrated augmentation has more room to help without distorting the conditional structure.
\vspace{0.1cm}

\noindent
Across both datasets, the key failure mode is \emph{mismatch between the synthetic data generating mechanism and the dependence/rarity structure of the real task}. This motivates evaluating augmentation methods not only on marginal realism, but on downstream performance under time-respecting splits and stress-regime diagnostics, as developed in the subsequent sections.

\section{Experimental Design}
\vspace{0.2cm}

\noindent
The objective of the experimental design is to systematically characterize the conditions under which synthetic augmentation improves or deteriorates out-of-sample performance in financial prediction tasks. We conduct our empirical analysis on two distinct data environments: a daily equity panel dataset and a high-frequency tick-by-tick options trade tape. These two settings differ substantially in frequency, sample size, signal-to-noise ratio, regime imbalance, and temporal dependence structure. This contrast allows us to evaluate whether the effectiveness of synthetic augmentation depends on the structural properties of the underlying learning problem rather than on dataset size alone.
\vspace{0.1cm}

\noindent
All experiments respect strict time ordering. For the daily equity panel, we employ a non-overlapping expanding walk-forward scheme with $K=5$ folds. The training window expands sequentially while the test window remains fixed in length, ensuring that each evaluation is fully out-of-sample and free of look-ahead bias. For the tick-by-tick dataset, we use chronological splits with strict separation between training and test blocks. No overlap between train and test periods is allowed in any configuration, and all preprocessing steps are computed within the training window only.
\vspace{0.1cm}

\noindent
We vary the augmentation ratio $\alpha = \frac{m}{m+n}$, where $n$ denotes the number of real training observations and $m$ the number of synthetic samples. The ratio $\alpha$ is evaluated over $\{0.25, 0.50, 0.75, 1.0\}$, allowing us to trace how performance evolves as the mixed training distribution 
\[
P_\alpha = (1-\alpha)P_{\text{real}} + \alpha P_{\text{synth}}
\]
moves progressively away from the empirical distribution. This grid permits us to observe when additional synthetic mass reduces estimation variance and when it instead induces distributional distortions.
\vspace{0.1cm}

\noindent
We compare multiple synthetic generators spanning increasing modeling flexibility. These include the stationary block bootstrap, Gaussian and Student-t copulas, variational autoencoders (VAE), denoising diffusion probabilistic models (DDPM), and TimeGAN. The bootstrap serves as a non-parametric benchmark that expands sample size without extrapolating beyond the empirical support. Copula-based methods preserve marginal distributions while modeling cross-sectional rank dependence. Deep generative models introduce latent representations capable of interpolation and limited extrapolation beyond observed samples. For generators that operate on static feature vectors (copula, VAE, DDPM), temporal flattening is applied to encode intra-window dynamics prior to generation, and synthetic windows are subsequently unflattened and aggregated to reconstruct time-series structure.
\vspace{0.1cm}

\noindent
For classification tasks, synthetic targets are generated as continuous outputs and subsequently mapped back to discrete labels using quantile matching so that the empirical class proportions are preserved exactly. This avoids artificial rebalancing effects that would otherwise confound the evaluation of generator quality.
\vspace{0.1cm}

\noindent
A central component of the design is the null-augmentation control. In addition to the baseline model trained on real data only and the synthetic-augmented model, we train a null model on real data plus a size-matched perturbation that preserves marginal structure but destroys predictive alignment, for example through label permutation or temporal shuffling. This construction isolates informational contribution from the mechanical effect of increasing sample size. Performance comparisons are therefore conducted both relative to the real-only baseline and relative to the null control.
\vspace{0.1cm}

\noindent
We evaluate augmentation across multiple predictive model classes in order to study interaction with model capacity and regularization. The model grid includes regularized linear classifiers (logistic regression with L1 and L2 penalties, Ridge classifier) and tree-based ensembles (random forest). Hyperparameters are selected within each training fold only. Because the learning algorithm is held fixed when comparing real, null, and synthetic training sets, any performance difference reflects the interaction between generator structure, model capacity, and the training distribution.
\vspace{0.1cm}

\noindent
In settings with pronounced regime imbalance, particularly in the tick dataset, where high-volatility observations are extremely rare, we additionally evaluate performance conditional on stressed states. By restricting the loss differential to rare regimes, we can detect localized improvements that may be masked in global averages.
\vspace{0.1cm}

\noindent
Statistical significance for all comparisons is assessed using the block permutation test introduced in Section \ref{sec:framework}. Permutation is conducted at the block level to preserve weak temporal dependence in the test set. For the walk-forward experiments, we report both fold-level significance and the proportion of folds rejecting the null hypothesis at the 5\% level. This ensures that conclusions about the usefulness of synthetic augmentation are not driven by isolated sample fluctuations.
\vspace{0.1cm}

\noindent
The design is therefore comparative and structural rather than benchmark-specific. By systematically varying generator type, augmentation ratio, and model capacity across two fundamentally different financial environments-one near-efficient daily panel and one highly non-Gaussian, regime-imbalanced high-frequency tape-we obtain a disciplined assessment of when synthetic augmentation provides incremental predictive information and when it fails.
\vspace{0.1cm}

\noindent
In addition to the empirical applications, the same experimental framework is applied in a fully controlled toy environment. There, the data-generating process is explicitly specified through a two-state Markov-switching model, allowing the true population distribution to be known. This setting permits a clean separation between variance reduction and distributional shift effects, since synthetic samples can be drawn either from the true DGP or from deliberately misspecified generators. By varying the augmentation ratio and the degree of structural mismatch, we obtain a transparent benchmark for interpreting the empirical findings in the real-data environments.
\section{Controlled Environment}
\vspace{0.1cm}

\noindent
This section provides the first empirical validation of the framework
in a fully controlled environment.
The objective is not realism,
but structural transparency:
since the data-generating process (DGP) is fully known,
we can cleanly separate variance reduction from population shift effects.
\vspace{0.1cm}

\noindent
We consider a two-state Markov-switching return process.
Let $S_t \in \{0,1\}$ denote the latent regime.
Conditional on $S_t=s$,

\begin{equation}
r_t \mid S_t=s \sim \mathcal N(\mu_s, \sigma_s^2).
\end{equation}

\noindent
The transition matrix is parameterized by $(p_{00}, p_{11})$,
which implies a stationary rare-state probability $\pi$.
Labels are defined using a forward return threshold:
\begin{equation}
Y_t = \mathbf 1 \{ r_{t+1} < \tau \}.
\end{equation}

\noindent
Features are noisy signals of the next-period return:
\begin{equation}
X_t = \beta r_{t+1} + \sigma_x \varepsilon_t,
\quad
\varepsilon_t \sim \mathcal N(0,1).
\end{equation}

\noindent
This construction yields a fully specified joint distribution
$(X_t, Y_t, S_t)$. We consider several synthetic generators:

\begin{itemize}
\item \textbf{Oracle}: new samples drawn from the exact same DGP.
\item \textbf{Oracle-Jitter}: small perturbations of structural parameters.
\item \textbf{Signal Mismatch}: inversion of $\beta$.
\item \textbf{Regime Mismatch}: distortion of the rare-state probability.
\item \textbf{Volatility Mismatch}: modification of $\sigma_1$.
\end{itemize}

\noindent
These generators induce controlled distributional shifts,
allowing us to study bias-dominant regimes. We then vary:
\vspace{0.1cm}

\noindent
\begin{itemize}
\item the mixture weight $\alpha = \frac{n}{m+n}$,
\item the rare-state probability $\pi$,
\item the type of synthetic generator.
\end{itemize}
\vspace{0.1cm}

\noindent
For each configuration,
we repeat the experiment $R$ times
and compute the median performance differential $\hat\delta$,
along with empirical rejection frequency.
We also compute a rare-regime restricted statistic
conditioning on $S_t=1$.
\vspace{0.1cm}

\subsection*{Results}
\vspace{0.1cm}

\paragraph{Oracle augmentation.}
When synthetic data is drawn from the true DGP,
performance improves monotonically in $\alpha$.
This behavior is consistent with pure variance reduction:
as the effective sample size increases,
estimation error decreases without inducing population shift.

\paragraph{Bias-dominant regimes.}
When structural mismatches are introduced
(e.g., sign-flipped signal),
performance deteriorates sharply for large $\alpha$.
This confirms the bias–variance decomposition:
as $\alpha$ increases,
the population objective shifts toward $P_{\mathrm{synth}}$,
and asymptotic bias dominates variance reduction.

\paragraph{Rare-regime effects.}
For regime-frequency distortions,
global performance may remain stable,
while rare-regime performance exhibits strong effects.
This highlights that synthetic augmentation
can selectively improve or degrade performance
in low-probability regions.
\vspace{0.1cm}

\noindent
These controlled experiments validate the theoretical framework.
We now turn to more realistic settings,
where the true DGP is unknown and synthetic generators
must be estimated from finite data.
\vspace{0.1cm}
\begin{table*}[!t]
\centering
\small
\setlength{\tabcolsep}{6pt}
\renewcommand{\arraystretch}{1.2}

\begin{tabular}{lcccc}
\toprule
Synthetic Design 
& Median $\hat\delta$ 
& Power (5\%) 
& Median $\hat\delta_{\text{rare}}$ 
& Rare Power (5\%) \\
\midrule

Oracle 
& 0.00132 
& 0.20 
& 0.00277 
& 0.70 \\

Signal mismatch ($\beta$ flip) 
& -0.00448 
& 0.00 
& -0.00189 
& 0.00 \\

Regime-frequency mismatch ($\pi$ high) 
& 0.00150 
& 0.15 
& 0.00414 
& 0.90 \\

Volatility mismatch ($\sigma_1$ low) 
& 0.00152 
& 0.15 
& 0.00363 
& 0.80 \\

\bottomrule
\end{tabular}

\caption{
Representative controlled-environment results at $\alpha=0.5$ and $\pi_{\text{rare}}=0.10$. 
Global and rare-regime performance differentials are reported alongside empirical rejection frequencies of the block-permutation test (5\% level). 
See Figure~\ref{fig:toy-heatmaps-2x2} for the full grid.
}

\label{tab:controlled-summary}

\end{table*}

\begin{figure*}[!t]
\centering

\begin{subfigure}[t]{0.49\textwidth}
\centering
\includegraphics[width=\linewidth]{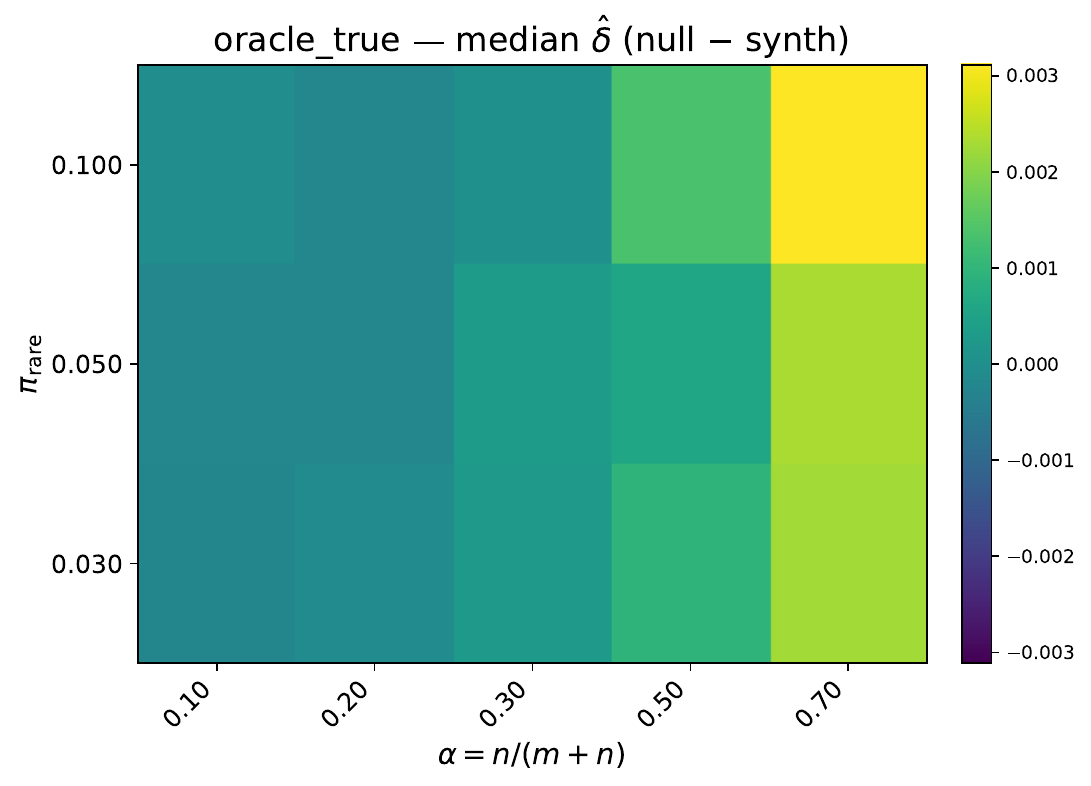}
\caption{Oracle (global) - median $\hat\delta$.}
\label{fig:hm-oracle}
\end{subfigure}\hfill
\begin{subfigure}[t]{0.49\textwidth}
\centering
\includegraphics[width=\linewidth]{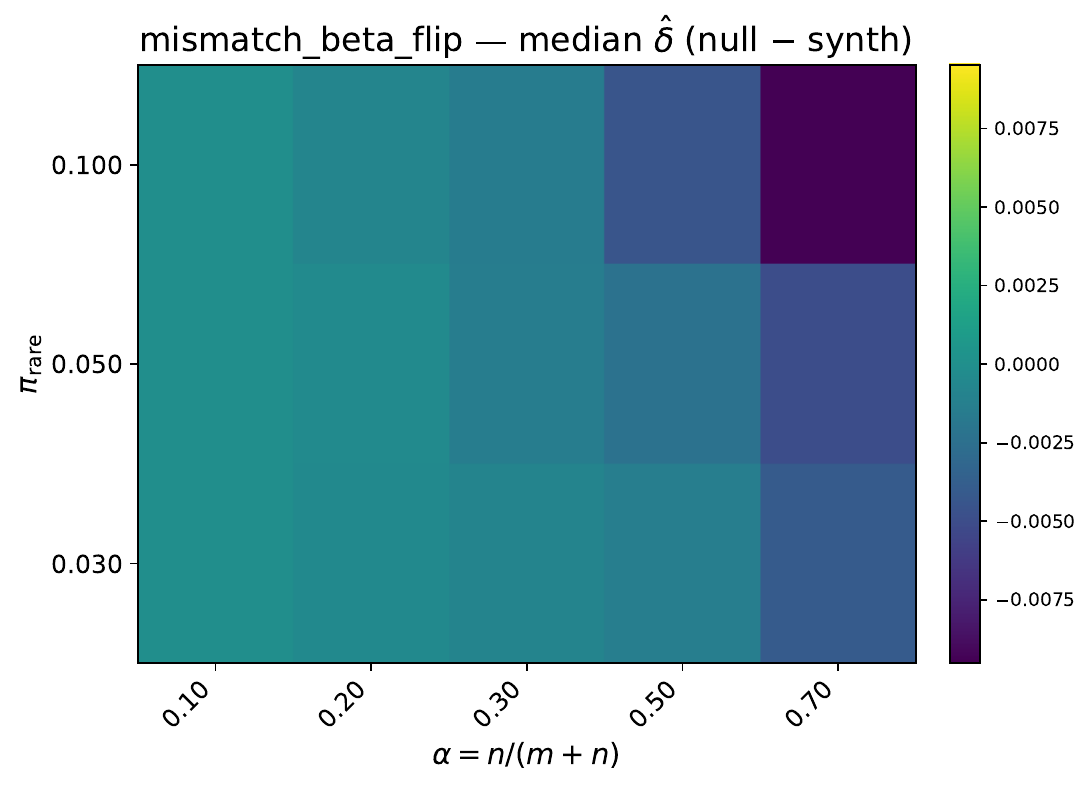}
\caption{Signal mismatch (global) - median $\hat\delta$.}
\label{fig:hm-beta}
\end{subfigure}

\vspace{0.35cm}

\begin{subfigure}[t]{0.49\textwidth}
\centering
\includegraphics[width=\linewidth]{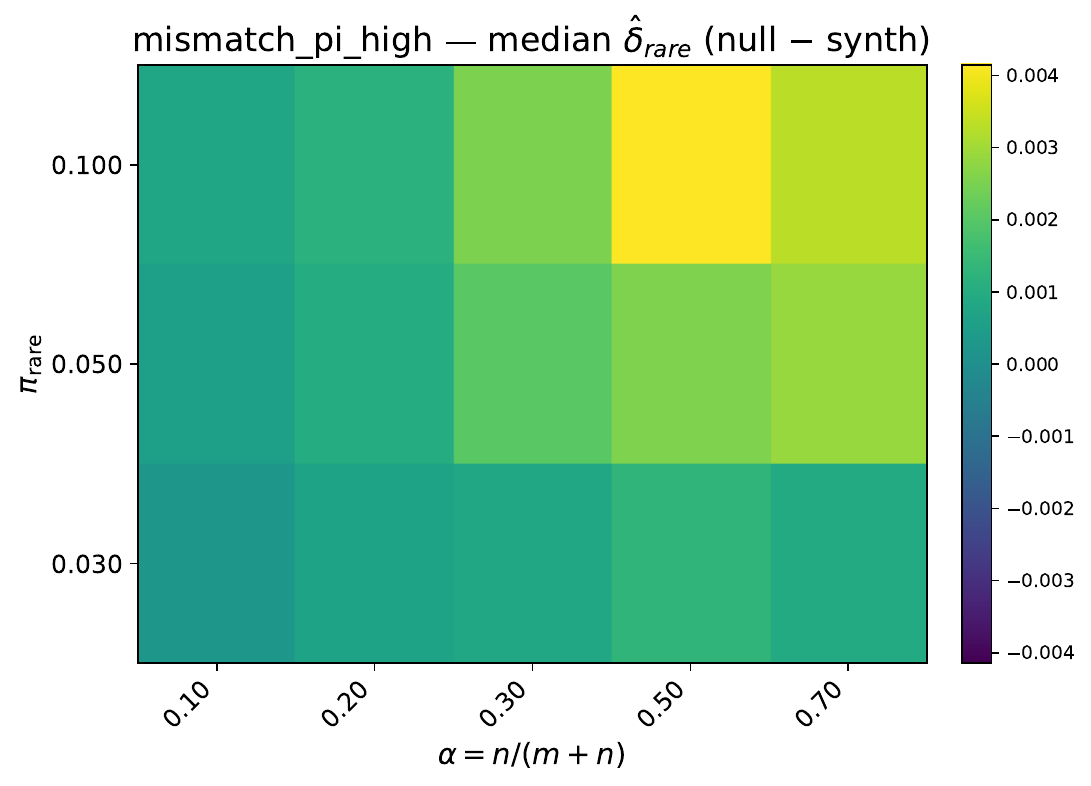}
\caption{Regime mismatch (rare) - median $\hat\delta_{\text{rare}}$ ($S_t=1$).}
\label{fig:hm-pi-rare}
\end{subfigure}\hfill
\begin{subfigure}[t]{0.49\textwidth}
\centering
\includegraphics[width=\linewidth]{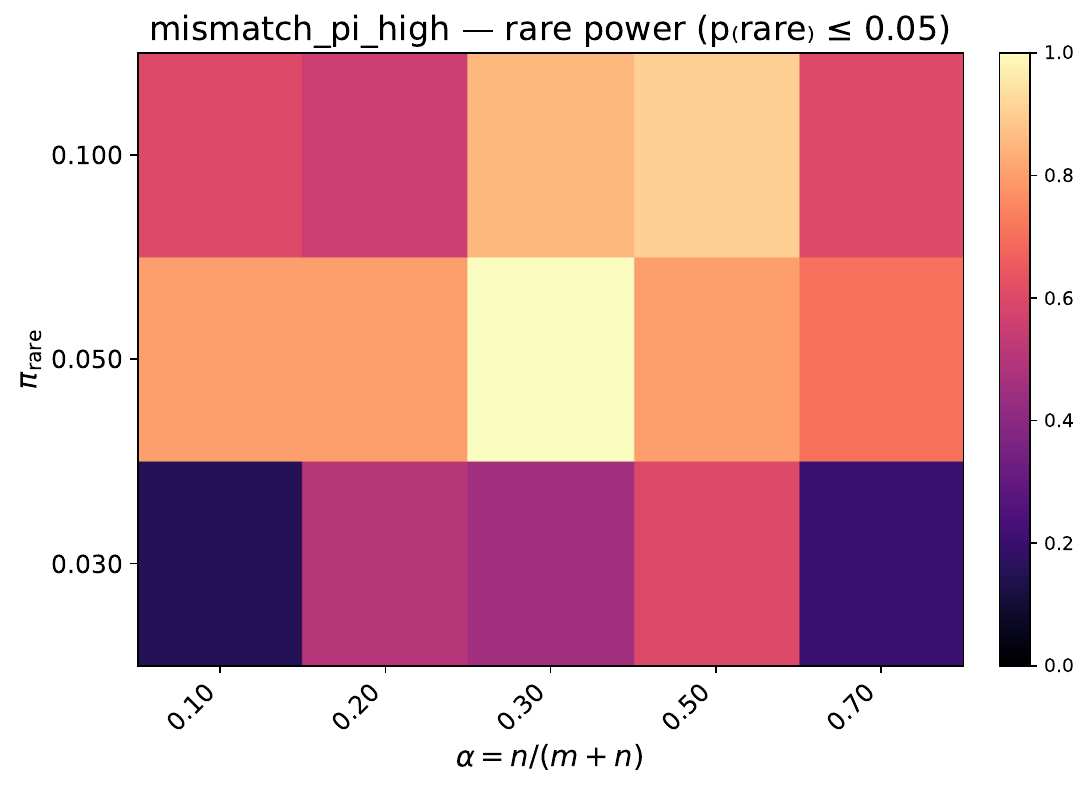}
\caption{Regime mismatch (rare) - power (block permutation).}
\label{fig:hm-pi-power}
\end{subfigure}

\caption{Controlled MS2 study. Heatmaps over $(\alpha,\pi_{\text{rare}})$ comparing oracle augmentation, signal mismatch, and regime mismatch with rare-regime evaluation.}
\label{fig:toy-heatmaps-2x2}
\end{figure*}

\noindent
Table~\ref{tab:controlled-summary} highlights the central mechanism revealed by the heatmaps in Figure~\ref{fig:toy-heatmaps-2x2}. 
When the synthetic generator is correctly specified (Oracle), augmentation improves performance both globally and within the rare regime. 
In contrast, a signal-level misspecification (sign flip in $\beta$) leads to a clear bias-dominant regime where performance deteriorates as $\alpha$ increases. 
Finally, regime-frequency and volatility mismatches primarily affect the rare regime, generating large local effects and high rejection power in rare-state tests.

\section{Synthetic Data Generators}\label{sec:generators}

\noindent
The choice of generative mechanism is central to the analysis.
Each generator defines a distinct $P_{\mathrm{synth}}$ and therefore
a distinct mixture distribution
$P_\alpha = (1-\alpha)P_{\mathrm{real}} + \alpha P_{\mathrm{synth}}$.
The bias--variance decomposition of $\Delta R(m,n)$ from
Section~\ref{sec:framework} makes clear that the alignment of
$P_{\mathrm{synth}}$ with $P_{\mathrm{test}}$ determines whether
augmentation is beneficial.
We select five generators that span a wide range along three axes:
capacity to extrapolate beyond observed support,
ability to model temporal dependence, and computational cost.
Table~\ref{tab:generators} provides an overview;
the subsections below give the intuition and key equations for each.

\subsection*{Block Bootstrap}\label{sec:gen:bootstrap}

\paragraph{Intuition.}
The stationary block bootstrap of \citep{kunsch1989} is the foundational
non-parametric baseline.
It introduces no distributional assumption and cannot produce observations
outside the support of the training data: $\mathrm{supp}(P_{\mathrm{synth}})
\subseteq \mathrm{supp}(P_{\mathrm{real}})$ by construction.
Any gain in $\hat\delta$ therefore reflects pure variance reduction from
sample-size expansion, not distributional extrapolation.
This makes it the natural benchmark against which learned generators are
compared.

\paragraph{Mechanism.}
A random start index $i \sim \mathcal{U}\{1,\ldots,T\}$ is drawn and a
block of consecutive observations of geometrically distributed length
\[
L \sim \mathrm{Geom}(1/\bar{b}), \qquad \bar{b} = \lceil n^{1/3} \rceil,
\]
is appended to the synthetic series.
The geometric distribution on block length, rather than a fixed length,
is the key device of \citep{kunsch1989}: it ensures the resampled series
is itself strictly stationary at block seams.
This is repeated until $n$ synthetic rows are assembled.
For static data the method reduces to standard i.i.d.\ resampling with
replacement.

\subsection*{TimeGAN}\label{sec:gen:timegan}

\paragraph{Intuition.}
TimeGAN \citep{yoon2019timegan} is the only generator in our suite
designed end-to-end for time series.
Its key contribution is conducting adversarial training in a latent space
first shaped by a supervised reconstruction objective, preventing the
discriminator from exploiting superficial spatial patterns while ignoring
temporal dynamics, a known failure mode of vanilla GANs on sequential
data.

\paragraph{Mechanism.}
Five GRU networks are trained in three sequential phases.
Let $\mathbf{x}_{1:S}$ be a training sequence of length $S$.
\vspace{0.1cm}

\noindent
\textit{Phase 1 - Embedding} (50\% of epochs):
an embedder $\mathcal{E}$ and recovery $\mathcal{R}$ are trained on
reconstruction, $\mathcal{L}_{\mathrm{rec}} = \sum_t \|x_t -
\mathcal{R}(\mathcal{E}(x_t))\|^2$, so the latent space mirrors data
geometry before adversarial training begins.
\vspace{0.1cm}

\noindent
\textit{Phase 2 - Supervised temporal} (25\% of epochs):
a generator $\mathcal{G}: \mathbf{z} \mapsto \tilde{\mathbf{h}}$ and
supervisor $\mathcal{S}: \tilde{\mathbf{h}}_t \mapsto \tilde{\mathbf{h}}_{t+1}$
are trained to reproduce step-to-step autoregressive dynamics in latent
space:
$\mathcal{L}_{\mathrm{sup}} = \sum_t \|\mathbf{h}_{t+1} -
\mathcal{S}(\tilde{\mathbf{h}}_t)\|^2$.
\vspace{0.1cm}

\noindent
\textit{Phase 3 - Joint adversarial} (25\% of epochs):
a discriminator $\mathcal{D}$ distinguishes real from fake latents under
the standard GAN objective, augmented by a moment-matching penalty
$\mathcal{L}_{\mathrm{mom}} = \|\mu_{\mathbf{h}} -
\mu_{\tilde{\mathbf{h}}}\|^2 + \|\sigma_{\mathbf{h}} -
\sigma_{\tilde{\mathbf{h}}}\|^2$.
At generation time, noise $\mathbf{z} \sim \mathcal{N}(\mathbf{0},
\mathbf{I})$ is passed through $\mathcal{G} \to \mathcal{S} \to
\mathcal{R}$ to produce synthetic sequences in the original space.

\subsection*{Denoising Diffusion Probabilistic Model}\label{sec:gen:ddpm}

\paragraph{Intuition.}
DDPM \citep{ho2020ddpm} defines $P_{\mathrm{synth}}$ through a learned
denoising operator rather than an encoder--decoder or a discriminator.
Its training objective is a simple mean-squared noise prediction, which
avoids the mode collapse and saddle-point dynamics of GAN training -
a meaningful stability advantage in the heavy-tailed, finite-sample
environments of our datasets.
Critically, the reverse diffusion process can generate samples outside
the convex hull of the training data, giving DDPM an extrapolation
capacity that the bootstrap lacks.

\paragraph{Mechanism.}
A forward Markov chain corrupts a real sample $\mathbf{x}_0$ over $T$
steps with a linear variance schedule $\beta_1 = 10^{-4} \to
\beta_T = 0.02$:
\[
\mathbf{x}_t
= \sqrt{\bar\alpha_t}\,\mathbf{x}_0
  + \sqrt{1-\bar\alpha_t}\,\boldsymbol\varepsilon,
\quad
\boldsymbol\varepsilon \sim \mathcal{N}(\mathbf{0},\mathbf{I}),
\]
where $\bar\alpha_t = \prod_{s=1}^t(1-\beta_s)$.
An MLP $\boldsymbol\varepsilon_\theta(\mathbf{x}_t,t)$ is trained to
predict $\boldsymbol\varepsilon$ via the simplified ELBO,
$\mathcal{L} = \mathbb{E}\|\boldsymbol\varepsilon -
\boldsymbol\varepsilon_\theta(\mathbf{x}_t,t)\|^2$.
Sampling starts from $\mathbf{x}_T \sim \mathcal{N}(\mathbf{0},
\mathbf{I})$ and applies the learned reverse step iteratively:
\[
\mathbf{x}_{t-1}
= \frac{1}{\sqrt{\alpha_t}}
  \!\left(
    \mathbf{x}_t
    - \frac{\beta_t}{\sqrt{1-\bar\alpha_t}}
      \boldsymbol\varepsilon_\theta(\mathbf{x}_t,t)
  \right)
+ \sqrt{\beta_t}\,\mathbf{z},
\]
\[
\quad \mathbf{z} \sim \mathcal{N}(\mathbf{0},\mathbf{I}).
\]
DDPM is deployed as a static generator on flattened-window inputs
(Section~\ref{sec:gen:flattening}).

\subsection*{Gaussian Copula}\label{sec:gen:copula}

\paragraph{Intuition.}
The Gaussian copula \citep{embrechts1999} is the only generator that
separates marginal distributions from the dependence structure by
construction, via Sklar's theorem.
This is well suited to financial data, where marginals are strongly
non-Gaussian (Section~\ref{sec:data}) while dependence can often be
captured by a flexible correlation model.
Requiring no gradient-based training, it is immune to the instabilities
affecting deep models in small-sample or heavy-tailed settings and
serves as an interpretable parametric reference point.

\paragraph{Mechanism.}
By Sklar's theorem, the joint distribution factors as
$F(x_1,\ldots,x_d) = C(F_1(x_1),\ldots,F_d(x_d))$,
where $C$ is a unique copula.
Each column is first mapped to its empirical uniform margin
$u_j = \hat F_j(x_j)$.
For the Gaussian family the copula is estimated by fitting a rank
correlation matrix $\hat{\boldsymbol\Sigma}$ on the normal-score
transforms $\Phi^{-1}(u_j)$.
New uniforms are sampled as
\[
\mathbf{U}_{\mathrm{new}}
= \Phi\!\left(\hat{\boldsymbol\Sigma}^{1/2}\boldsymbol\xi\right),
\quad \boldsymbol\xi \sim \mathcal{N}(\mathbf{0},\mathbf{I}),
\]
then each column is back-transformed via the empirical quantile function
$\hat F_j^{-1}$, guaranteeing exact marginal replication.
The Student-$t$ copula variant introduces symmetric tail dependence
via its degrees-of-freedom parameter $\nu$, relevant for joint
crash dynamics in the option datasets.

\subsection*{Variational Autoencoder}\label{sec:gen:vae}

\paragraph{Intuition.}
The VAE \citep{kingma2013vae} learns a smooth, compressed latent
representation of $P_{\mathrm{real}}$, regularised by a KL divergence
toward a standard Gaussian prior.
This regularised latent space enables interpolation and mild extrapolation
between observed data points - qualitatively different from the
bootstrap (resampling only) and the copula (no latent compression).
Operating in a bottleneck of dimension $d \ll D \times W$, the VAE also
implicitly regularises generation in the high-dimensional flattened
feature space, which is particularly beneficial for the Volatility
Surface dataset.

\paragraph{Mechanism.}
An encoder maps $\mathbf{x}$ to the parameters of a Gaussian posterior,
$q_\phi(\mathbf{z}|\mathbf{x}) = \mathcal{N}(\boldsymbol\mu_z,
\mathrm{diag}(\boldsymbol\sigma_z^2))$;
a decoder reconstructs $p_\theta(\mathbf{x}|\mathbf{z})$.
Training maximises the ELBO:
\[
\mathcal{L}_{\mathrm{VAE}}
= \mathbb{E}_{q_\phi}\!\bigl[\log p_\theta(\mathbf{x}|\mathbf{z})\bigr]
  - \beta\,\mathrm{KL}\!\bigl(q_\phi(\mathbf{z}|\mathbf{x})
    \,\|\,\mathcal{N}(\mathbf{0},\mathbf{I})\bigr).
\]
The reparameterisation trick
$\mathbf{z} = \boldsymbol\mu_z + \boldsymbol\sigma_z \odot \boldsymbol\varepsilon$,
$\boldsymbol\varepsilon \sim \mathcal{N}(\mathbf{0},\mathbf{I})$,
makes sampling differentiable.
At generation time, $\mathbf{z} \sim \mathcal{N}(\mathbf{0},\mathbf{I})$
is passed through the decoder only.
The KL weight $\beta \geq 1$ ($\beta$-VAE) allows trading reconstruction
fidelity for latent disentanglement.

\subsection*{Temporal Flattening for Static
            Generators}\label{sec:gen:flattening}

The VAE, DDPM, and Copula treat every row as an independent draw:
temporal ordering is invisible to them.
Temporal flattening encodes intra-window dynamics as ordinary
cross-feature correlation, allowing any static generator to be applied
to time series without modification.

\paragraph{Flattening.}
A time series of $T$ rows and $D$ columns is sliced into windows of
length $W$ with stride $s \leq W$, yielding
\[
n_w = \left\lfloor \frac{T - W}{s} \right\rfloor + 1
\]
windows.
Each window is concatenated into a single row of $D \times W$ features:
\[
\begin{bmatrix}
x^{(1)}_0 & \cdots & x^{(D)}_0 \\
\vdots     &        & \vdots     \\
x^{(1)}_{W-1} & \cdots & x^{(D)}_{W-1}
\end{bmatrix}
\]
\[
\;\longrightarrow\;
\bigl[
  x^{(1)}_0,\ldots,x^{(D)}_0,\;
  \ldots,\;
  x^{(1)}_{W-1},\ldots,x^{(D)}_{W-1}
\bigr].
\]
The static generator thus captures the full intra-window second-order
structure through a single joint covariance matrix over $D \times W$
features.

\paragraph{Stride and reconstruction.}
Setting $s < W$ introduces overlapping windows, growing $n_w$ by a
factor $\approx W/s$ at the cost of inducing correlation between
consecutive training samples---useful when real data is scarce.
After generation, $N$ synthetic windows are unflattened into a series
of $T_{\mathrm{out}} = (N-1)s + W$ rows.
In the overlapping case each time step $\tau$ is covered by $c_\tau$
independent windows; the default aggregation is their mean,
\[
\tilde{x}_\tau
= \frac{1}{c_\tau}
  \sum_{k:\,ks \leq \tau < ks+W}
  \tilde{x}^{(k)}_{\tau-ks},
\]
which reduces per-step variance by a factor $c_\tau$ (a median option
is also available for robustness).
Only intra-window dependence is captured; dynamics spanning more than
$W$ lags and inter-window continuity are not modelled-for those,
TimeGAN or the block bootstrap are preferable.

\subsection*{Classification Post-Processing}\label{sec:gen:classif}

All generators produce continuous outputs, including the target column.
For classification tasks (directional labels $\{-1,0,+1\}$ on the tick
tape) these must be mapped to discrete classes.
Naive rounding fails whenever the generator shifts or rescales the
target marginal, distorting the class balance of $P_\alpha$ and
introducing a non-vanishing population shift in the $\Delta R$
decomposition.

\paragraph{Quantile matching.}
Let $p_k = \#\{y_i = c_k\}/N$ be the empirical class proportion in the
real data.
Each synthetic target value is mapped to a uniform quantile
$q_i = (\mathrm{rank}(i)+0.5)/N_s$, then assigned label $c_k$ if
\[
q_i \;\in\; \Bigl[\sum_{j<k} p_j,\;\sum_{j \leq k} p_j\Bigr).
\]
The synthetic class distribution is thereby \emph{exactly} equal to the
real one by construction, the rank ordering of generated values is
preserved, and the procedure extends to any number of classes in
$O(N_s \log N_s)$.

\subsection*{Generator Selection Summary}\label{sec:gen:summary}

\noindent
The five generators span the space from \emph{no learned structure}
(bootstrap) to \emph{full sequence learning} (TimeGAN).
The three static generators (Copula, VAE, DDPM) are made
time-series-aware through temporal flattening, capturing intra-window
dynamics as cross-feature correlation.
This design allows us to disentangle, in the subsequent experiments,
the contributions of (i) pure sample-size expansion, (ii)
distributional extrapolation, and (iii) temporal representation
learning to the risk differential $\Delta R$.

\end{multicols}
\begin{table*}[ht]
\centering
\caption{%
  Comparison of the five synthetic data generators along dimensions
  relevant to the bias--variance decomposition of $\Delta R$.
  \emph{Population shift risk} reflects whether the generator can
  introduce systematic misspecification ($P_{\mathrm{synth}} \neq
  P_{\mathrm{real}}$).
  \emph{Variance reduction mechanism} indicates how the generator
  reduces estimation error.
  \emph{Rare-regime coverage} indicates whether the generator can
  oversample the tail region $\mathcal{A}$.
  \emph{Temporal structure} reflects the depth of sequential dependence
  modelled.
  \emph{Small-$n$ stability} assesses reliability when real training
  data is scarce.
}
\label{tab:generators}
\small
\renewcommand{\arraystretch}{1.4}
\begin{tabular}{@{}p{2.1cm}p{2.6cm}p{2.8cm}p{2.4cm}p{2.6cm}p{2.1cm}@{}}
\toprule
\textbf{Generator}
  & \textbf{Population shift risk}
  & \textbf{Variance reduction mechanism}
  & \textbf{Rare-regime coverage}
  & \textbf{Temporal structure}
  & \textbf{Small-$n$ stability} \\
\midrule
Block Bootstrap
  & None by construction ($P_{\mathrm{synth}} \subseteq P_{\mathrm{real}}$)
  & Sample-size expansion only
  & No, inherits scarcity of real data
  & Block-wise autocorrelation (up to $\bar{b}$ lags)
  & Very high \\[4pt]
TimeGAN
  & Moderate, GAN training can mis-specify dynamics
  & Latent-space interpolation + sequence generation
  & Partial, can generate unseen sequences but not extrapolate marginals
  & Full sequence via GRU (autoregressive latent dynamics)
  & Low, unstable on small datasets \\[4pt]
DDPM
  & Low, stable training, smooth reverse diffusion
  & Learned data manifold + diffusion noise injection
  & Yes, reverse process generates samples beyond training support
  & Intra-window via flattening ($\leq W$ lags)
  & Moderate, needs sufficient epochs \\[4pt]
Copula
  & Low, marginals exactly preserved by construction
  & Exact marginal replication + flexible rank dependence
  & Partial, tail extrapolation via quantile inversion
  & Intra-window via flattening ($\leq W$ lags)
  & Very high, no gradient optimisation \\[4pt]
VAE
  & Low to moderate, KL regularisation may smooth tails
  & Latent interpolation in compressed bottleneck
  & Partial, smooth latent prior may under-represent extremes
  & Intra-window via flattening ($\leq W$ lags)
  & Moderate, sensitive to latent dim choice \\
\bottomrule
\end{tabular}
\end{table*}
\begin{multicols}{2}

\newpage

\section{Empirical Results: Daily Equity Panel}\label{sec:equity}

\paragraph{Experimental setup.}
We use a non-overlapping walk-forward scheme with $K=5$ folds:
training windows expand from 820 to 4{,}100 observations while the test window
is fixed at 820 observations, preserving temporal ordering throughout.
Twelve synthetic generators are evaluated: iid bootstrap, stationary block bootstrap,
Gaussian copula, Student-$t$ copula ($\nu=4$), Gumbel copula,
temporally flattened copula variants, VAE, DDPM and TimeGAN, at augmentation ratios
$\alpha \in \{0.25, 0.5, 0.75, 1.0\}$.
All comparisons use the size-matched null augmentation and block permutation test
described in Section~\ref{sec:framework}.
Each heatmap shows the best performance differential $\hat\delta$ per (generator, model) pair,
optimised over $\alpha$, with cells marked if $p<0.05$ in the majority of folds.

\subsection*{Classification Task}\label{sec:equity:clf}

The directional classification task is the most challenging.
The baseline achieves AUC $\approx 0.518$ and accuracy $\approx 51\%$,
barely above random, consistent with near-efficiency of large-cap U.S.\ equities at daily frequency.

\begin{figure}[H]
\centering
\includegraphics[width=0.9\linewidth]{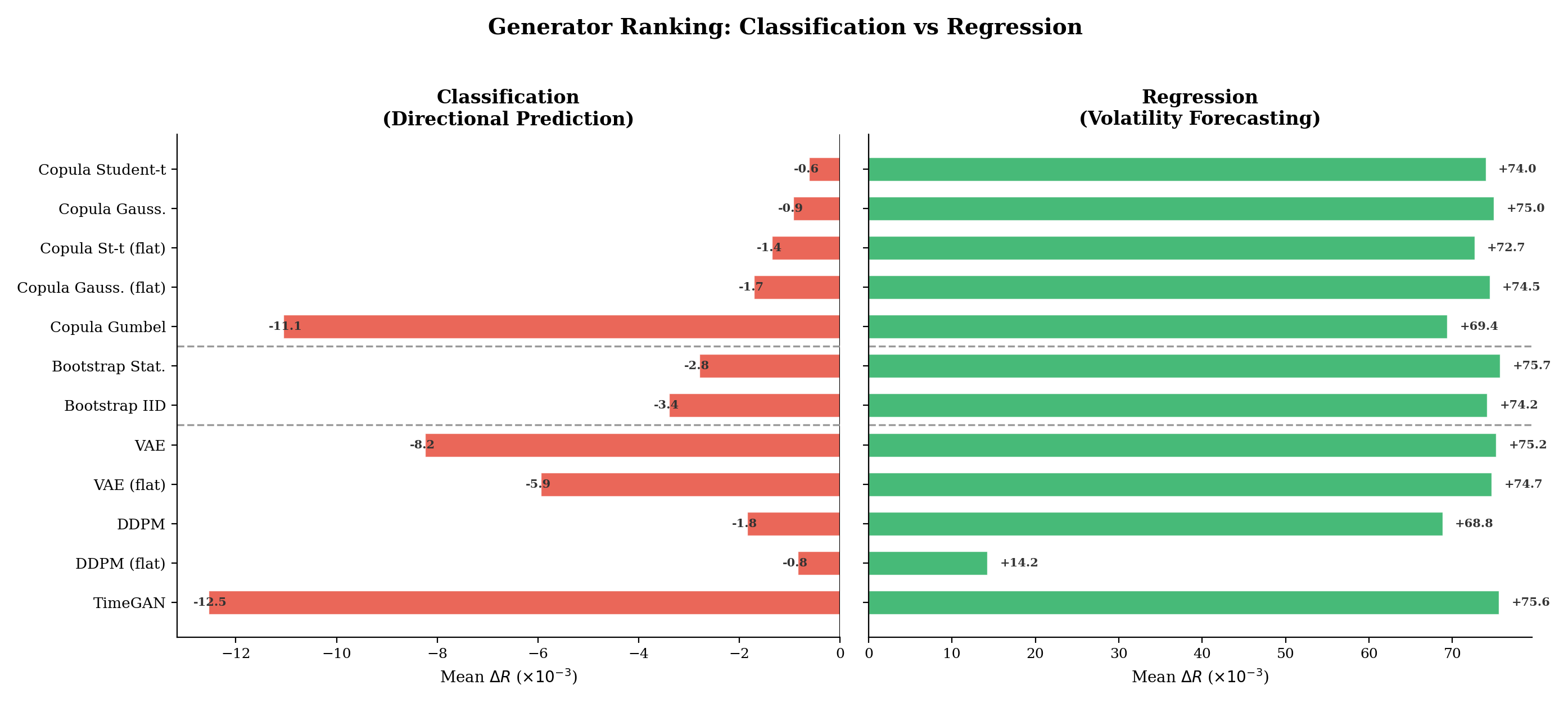}
\caption{%
  \textbf{Generator ranking: directional classification (left)
  vs.\ volatility forecasting (right)}, averaged over all models and ratios.
  Every generator is net negative for classification (all bars red)
  and strongly positive for volatility forecasting (all bars green),
  with magnitudes 50--100$\times$ larger.
  The contrast illustrates how baseline signal strength governs
  whether augmentation can be beneficial.%
}
\label{fig:cross_task}
\end{figure}

\paragraph{Augmentation hurts more than it helps.}
Figure~\ref{fig:cross_task} (left) provides the clearest summary:
averaged over all augmentation ratios, \emph{every generator} produces a negative
mean differential for classification.
The least harmful are the copula families
(Copula Student-$t$: $-0.6\times10^{-3}$; Copula Gaussian: $-0.9\times10^{-3}$),
while deep generative models are the most destructive
(TimeGAN: $-12.5\times10^{-3}$; VAE: $-8.2\times10^{-3}$).
\vspace{0.1cm}

\begin{figure}[H]
\centering
\includegraphics[width=0.9\linewidth]{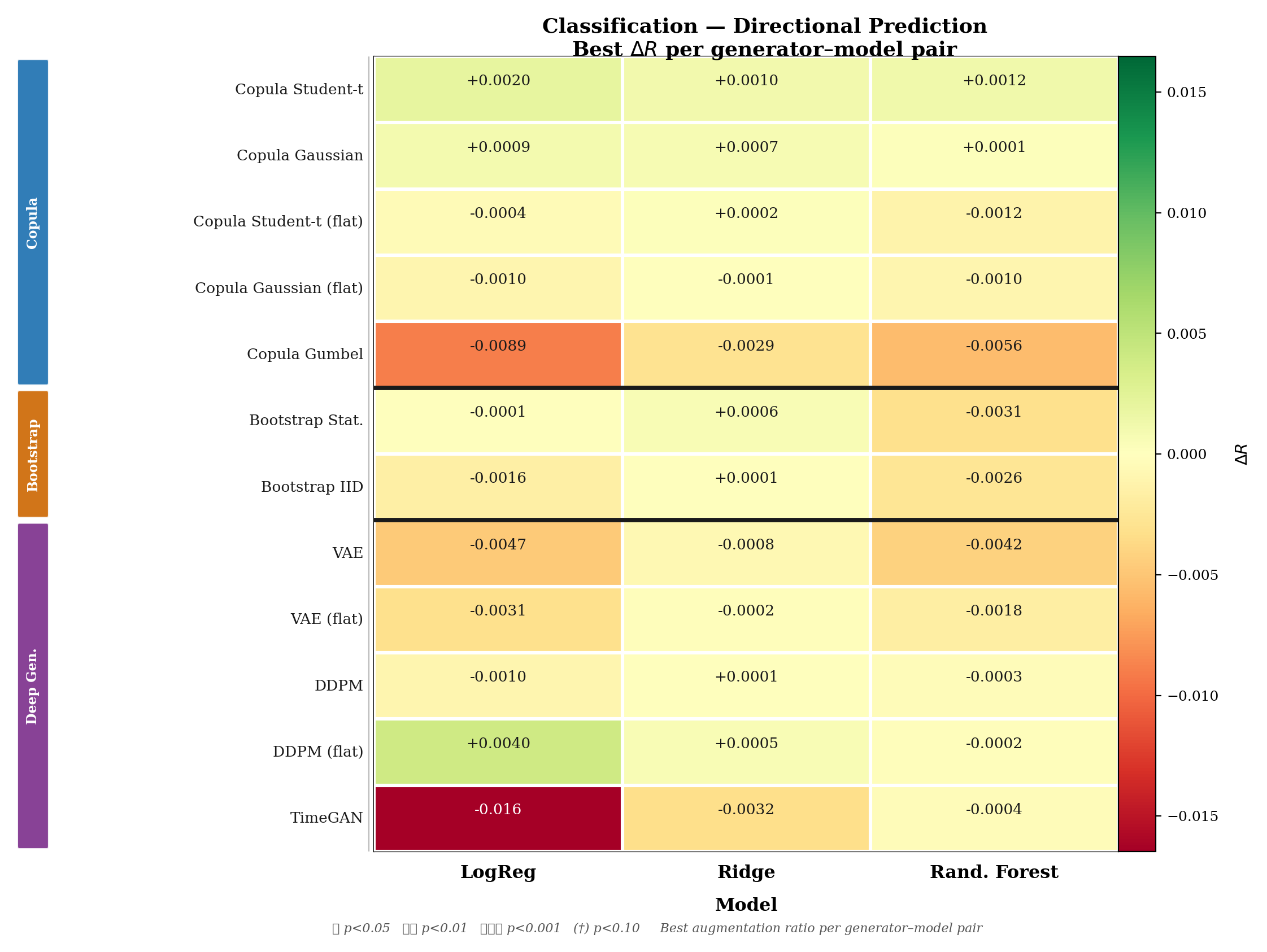}
\caption{%
  \textbf{Classification: best $\hat\delta$ per (generator, model) pair},
  optimised over augmentation ratios.
  Values are on the order of $10^{-3}$; only the Copula Student-$t$ row is
  consistently non-negative. TimeGAN and VAE are the worst performers.
  Significance markers appear in very few cells, reflecting near-zero
  statistical power across the board.%
}
\label{fig:clf_heatmap}
\end{figure}

\noindent
Even in the best-case view of Figure~\ref{fig:clf_heatmap} (values optimised over ratios),
magnitudes remain on the order of $10^{-3}$.
Only the Copula Student-$t$ row is consistently non-negative
(LogReg: $+0.0020$, Ridge: $+0.0010$, RF: $+0.0012$);
all other generators either break even or degrade performance.
Statistical significance is nearly absent: the maximum rejection rate
is 27\% (Copula Student-$t$ at $\alpha=0.25$), with every other cell at 0\%.

\paragraph{Ratio effect.}
Figure~\ref{fig:ratio_effect} (left) traces the ratio sweep for Copula Student-$t$.
The differential is marginally positive for LogReg at $\alpha=0.25$,
then all models fall into negative territory from $\alpha=0.50$ onward.
The near-zero baseline signal leaves almost no room for variance reduction:
any misspecification of $P_{\mathrm{synth}}$ shifts the population objective
away from $P_{\mathrm{real}}$ and degrades the predictor.
Synthetic augmentation provides no reliable benefit for directional equity prediction
at daily frequency.

\end{multicols}

\begin{multicols}{2}

\subsection*{Volatility Forecasting Task}\label{sec:equity:reg}

The regression task is one-step-ahead forecasting of realized volatility
(\texttt{vol\_5d}$_{t+1}$), substantially more predictable than returns
due to volatility clustering.
The baseline RF achieves $R^2 = 0.721$; Ridge/Linear reach $R^2 = 0.638$.

\begin{figure}[H]
\centering
\includegraphics[width=\linewidth]{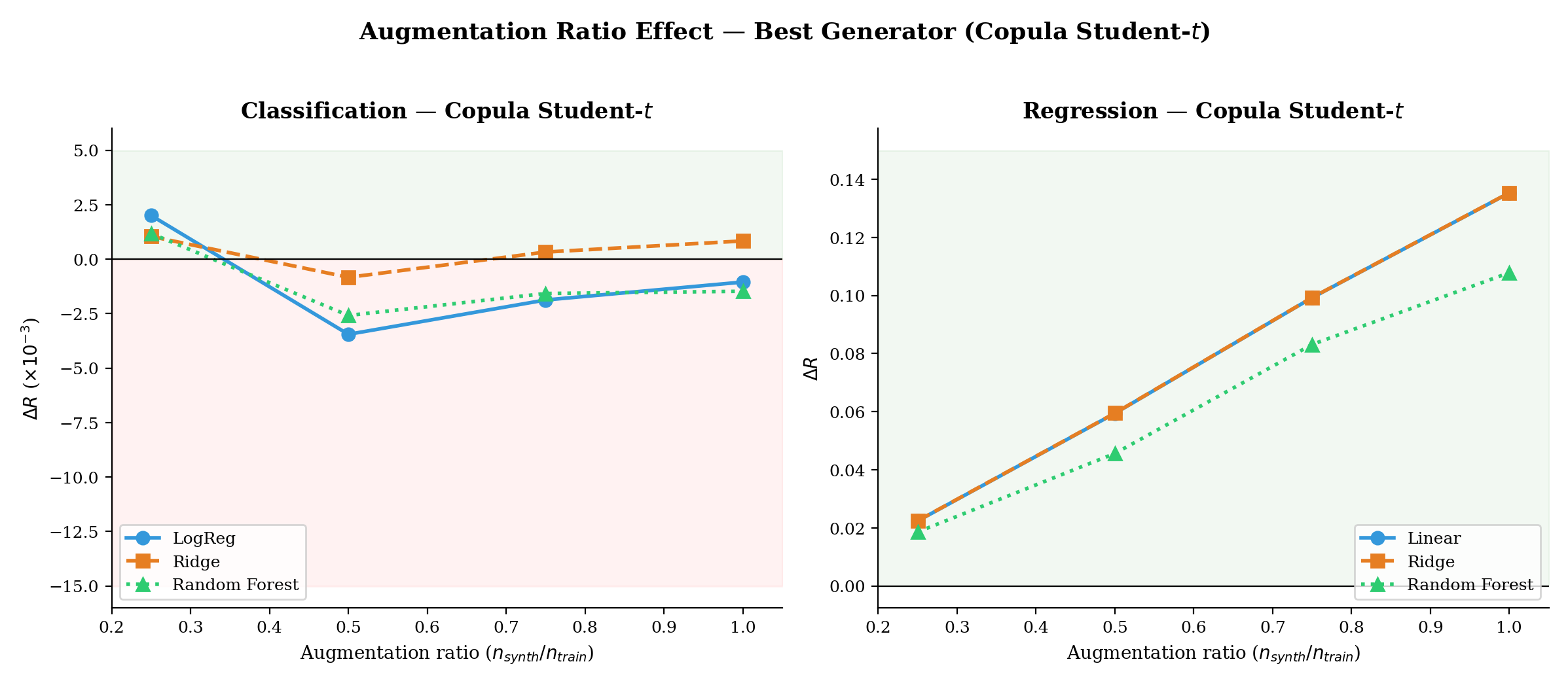}
\caption{%
  \textbf{Augmentation ratio effect for Copula Student-$t$}, averaged over models.
  \emph{Left (classification):} starts marginally positive at $\alpha=0.25$,
  then all models drop into negative territory from $\alpha=0.50$ onward
  (pink background dominates).
  \emph{Right (volatility forecasting):} positive across all models and
  increases monotonically with ratio (entirely green background).%
}
\label{fig:ratio_effect}
\end{figure}

\begin{figure}[H]
\centering
\includegraphics[width=\linewidth]{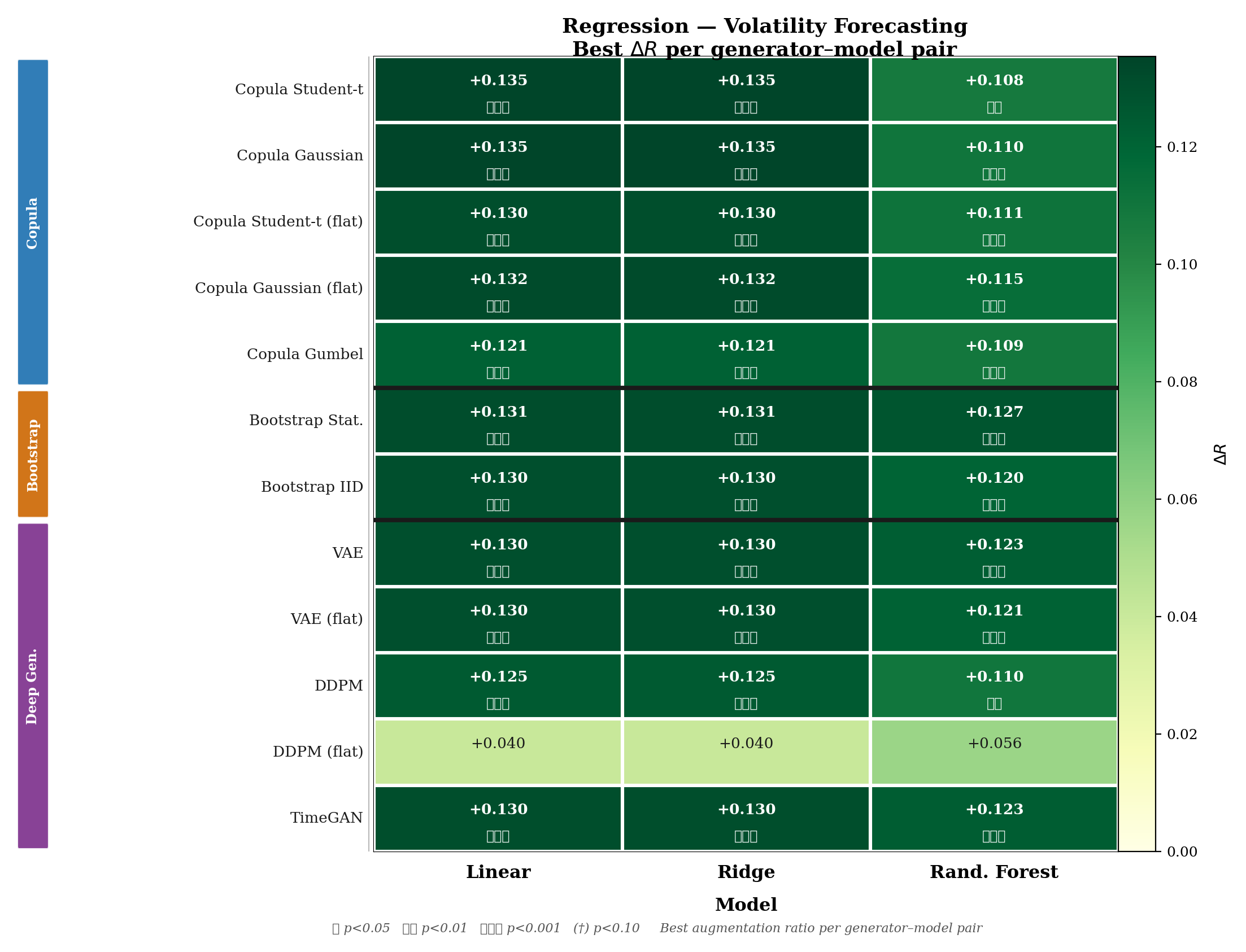}
\caption{%
  \textbf{Volatility forecasting: best $\hat\delta$ per (generator, model) pair.}
  Every cell is positive and triple-significant ($p<0.001$),
  in stark contrast to the classification heatmap.
  Copula Student-$t$ and Gaussian achieve $\hat\delta=+0.135$ for linear models;
  DDPM (flat) is the weakest ($+0.040$--$+0.056$).%
}
\label{fig:reg_heatmap}
\end{figure}

\paragraph{Augmentation effect.}
Figure~\ref{fig:reg_heatmap} presents a sharp contrast with the classification heatmap:
\emph{every cell is dark green} and triple-significant ($p<0.001$),
regardless of generator or model.
Top $\hat\delta$ values reach $+0.135$ for copula families paired with linear models,
and $+0.108$--$+0.127$ for RF.
The one exception is DDPM (flat), markedly weaker ($+0.040$--$+0.056$), likely
because temporal flattening distorts the local persistence structure of volatility.
Figure~\ref{fig:ratio_effect} (right) confirms that $\hat\delta$ increases
monotonically with $\alpha$ across all model types,
the mirror image of the classification panel.
\vspace{0.1cm}

\noindent
The practically relevant gain, comparing synthetic-augmented to \emph{unaugmented}
training, is more modest: $\Delta R^2 \approx +0.011$ for Copula Gaussian + Ridge
(Table~\ref{tab:reg_summary}), and essentially zero for RF.
Synthetic data adds signal for linear models by smoothing the dependence structure,
but offers little to models that already exploit nonlinear volatility dynamics.

\end{multicols}
\begin{table}[H]
\centering
\caption{%
  Volatility forecasting: top configurations ($\alpha=1.0$, 5-fold mean).
  $\hat\delta$: loss differential vs.\ null; $\Delta R^2$: gain vs.\ unaugmented baseline.%
}\label{tab:reg_summary}
\small
\setlength{\tabcolsep}{4pt}
\begin{tabular}{@{}llrrrr@{}}
\toprule
Generator & Model & $\hat\delta$ & $p$ & $R^2_{\text{base}}$ & $\Delta R^2$ \\
\midrule
Copula $t_4$  & Linear  & $+0.137$ & $<$0.001 & 0.638 & $+0.011$ \\
Copula Gauss  & Linear  & $+0.137$ & $<$0.001 & 0.638 & $+0.011$ \\
Copula $t_4$  & Ridge   & $+0.137$ & $<$0.001 & 0.638 & $+0.011$ \\
Copula Gauss  & Ridge   & $+0.137$ & $<$0.001 & 0.638 & $+0.011$ \\
Boot.\ stat.  & RF      & $+0.140$ & $<$0.001 & 0.721 & $-0.001$ \\
\midrule
DDPM (flat)   & Linear  & $+0.040$ & $<$0.001 & 0.638 & $+0.001$ \\
\bottomrule
\end{tabular}
\end{table}

\begin{multicols}{2}

\subsection*{Rare Volatility-Spike Prediction}\label{sec:equity:rare}

\begin{figure}[H]
\centering
\includegraphics[width=0.85\linewidth]{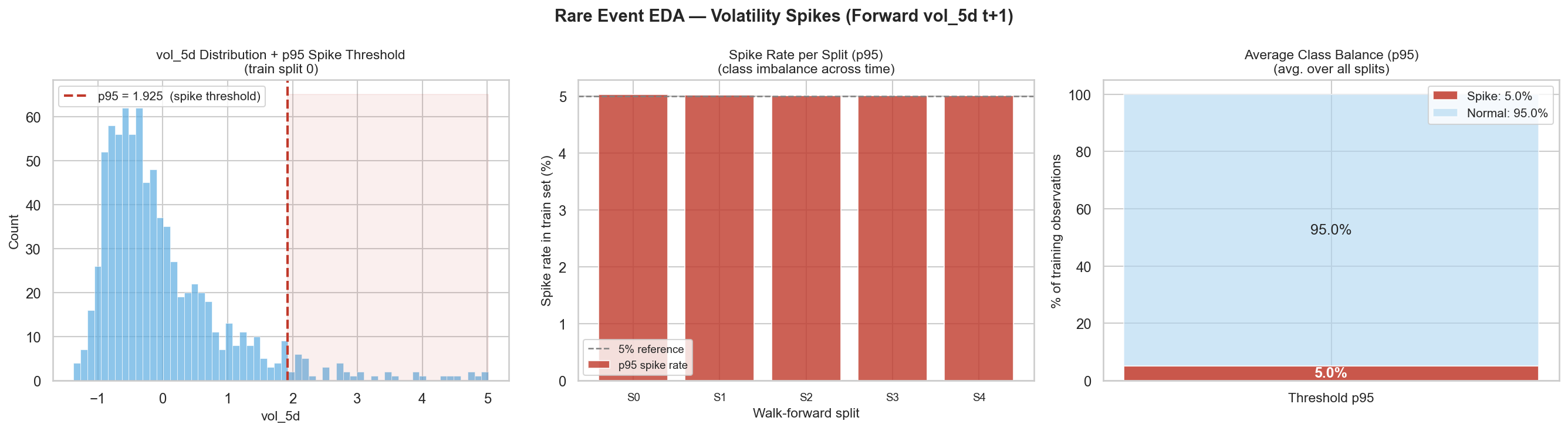}
\caption{%
  \textbf{Rare events EDA.}
  \emph{Left:} \texttt{vol\_5d} distribution with the p95 spike threshold
  (dashed at $z\approx1.925$).
  \emph{Centre:} spike rate per walk-forward fold, stable at $\approx5\%$.
  \emph{Right:} overall class balance confirming the 5\%/95\% imbalance.%
}
\label{fig:rare_eda}
\end{figure}

A \emph{volatility spike} is defined as \texttt{vol\_5d}$_{t+1}$
exceeding the in-sample 95th percentile (z-scored threshold $\approx 1.925$),
yielding a stable 5\%/95\% imbalance across all folds (Figure~\ref{fig:rare_eda}).
Average Precision (AP) is the primary metric.
We evaluate GBM, XGBoost, balanced RF (RF$_{\text{bal}}$),
and balanced logistic regression (LogReg$_{\text{bal}}$),
under two augmentation strategies:
\emph{vanilla} (unconditional, preserves 5\% spike frequency)
and \emph{targeted} (oversamples the spike regime exclusively,
in the spirit of SMOTE applied to a generative model).

\end{multicols}
\begin{multicols}{2}

\begin{figure}[H]
\centering
\includegraphics[width=\linewidth]{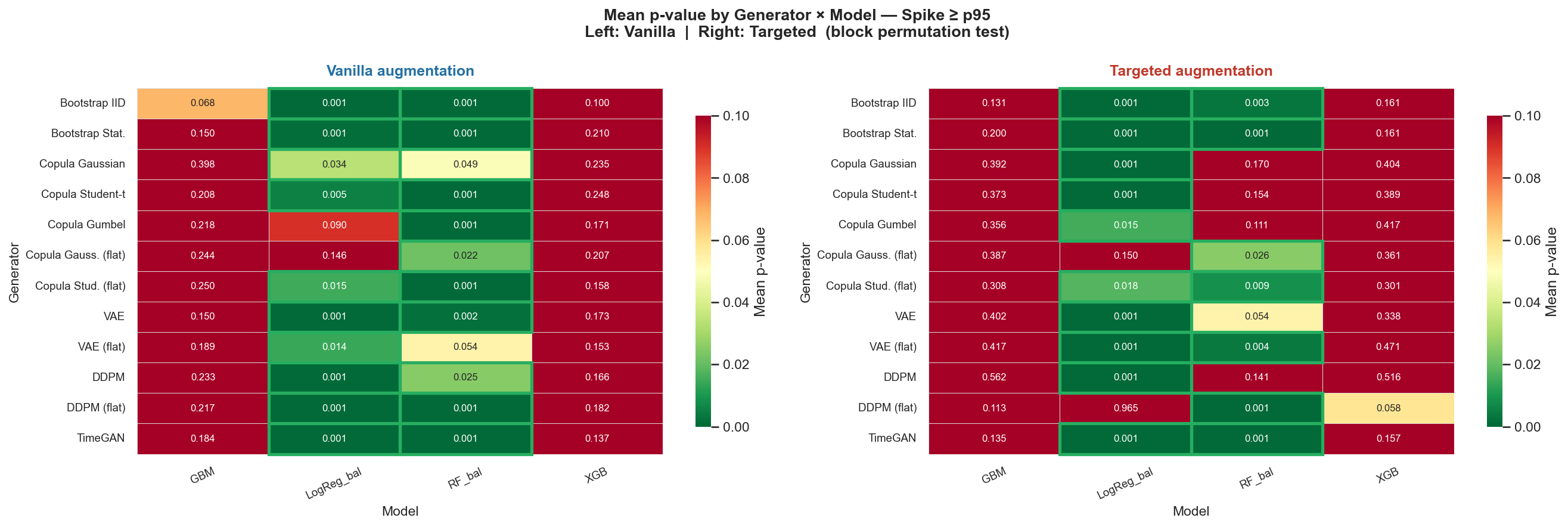}
\caption{%
  \textbf{Mean permutation-test p-value by generator $\times$ model
  (volatility spike $\geq$ p95).}
  \emph{Left (vanilla):} LogReg$_{\text{bal}}$ achieves the lowest p-values
  (dark green); GBM and XGB are much less significant.
  \emph{Right (targeted):} LogReg$_{\text{bal}}$ dominates even more strongly,
  while GBM and XGB remain moderately significant.
  Despite low p-values, LogReg$_{\text{bal}}$ configurations yield no AP
  improvement; configurations with genuine AP gains (DDPM flat + XGB,
  TimeGAN + XGB) appear mid-range here.%
}
\label{fig:rare_pvalues}
\end{figure}

\paragraph{Permutation test p-values.}
Figure~\ref{fig:rare_pvalues} shows mean p-values per (generator, model) pair
under vanilla (left) and targeted (right) augmentation.
A clear pattern emerges: \emph{LogReg$_{\text{bal}}$ dominates in statistical
significance}, with p-values at or near zero for most generators under both strategies,
and even more pronounced under targeting.
GBM and XGB remain moderately significant ($p \approx 0.03$--$0.5$);
RF$_{\text{bal}}$ is intermediate.

\paragraph{Permutation test vs.\ AP: a structural divergence.}
The most statistically significant model is \emph{not} the best at improving
rare-event detection, a structural divergence.
DDPM (flat) + XGB achieves the largest targeted AP gain ($+0.125$)
and TimeGAN + XGB follows ($+0.108$), yet both carry only moderate p-values.
LogReg$_{\text{bal}}$, dominant in significance, yields no AP improvement.
\vspace{0.1cm}

\noindent
The explanation is structural: LogReg$_{\text{bal}}$ minimises
a class-reweighted log-loss that already artificially up-weights minority examples;
any augmentation, even a label-permuted null, strongly disturbs this calibration,
producing a large, easily significant $\hat\delta$ that is irrelevant to spike-detection
accuracy.
GBM and XGB are genuinely helped by targeted oversampling for AP,
but the permutation benchmark penalises the distributional shift that makes
targeting effective, suppressing their significance.
\vspace{0.1cm}

\begin{figure}[H]
\centering
\includegraphics[width=\linewidth]{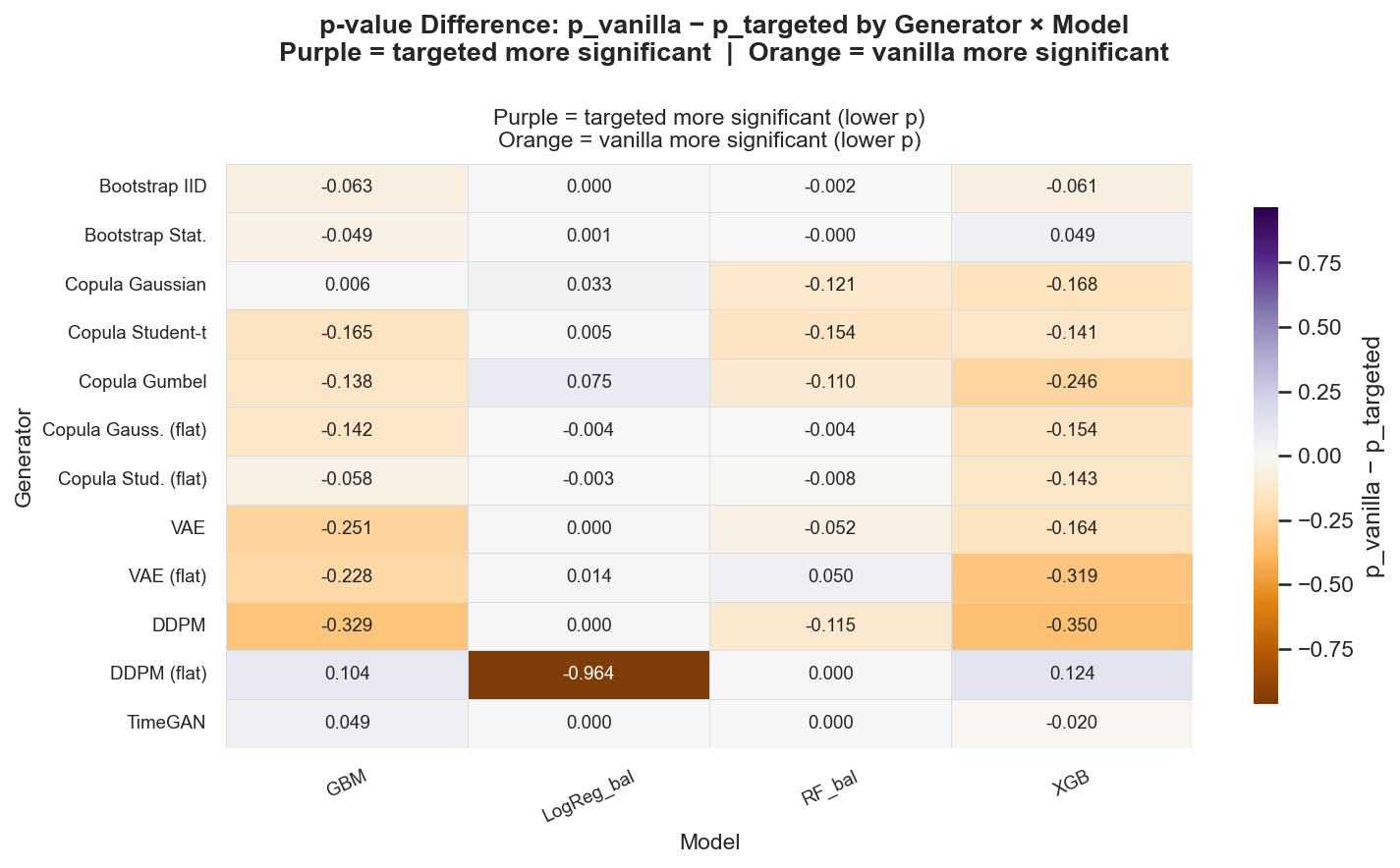}
\caption{%
  \textbf{P-value difference: $p_{\mathrm{vanilla}} - p_{\mathrm{targeted}}$.}
  Orange = vanilla more significant; purple = targeted more significant.
  The predominantly orange map confirms that targeted augmentation
  \emph{raises} permutation p-values relative to vanilla,
  even for configurations with the best AP gains.
  DDPM (flat) + LogReg$_{\text{bal}}$ is the extreme case ($\Delta p = -0.964$).%
}
\label{fig:rare_pvdiff}
\end{figure}

\noindent
Figure~\ref{fig:rare_pvdiff} maps $p_{\mathrm{vanilla}} - p_{\mathrm{targeted}}$
across all pairs.
The predominantly orange heatmap confirms that targeting a rare regime
\emph{systematically raises} permutation-test p-values relative to unconditional
augmentation, even when it improves the domain metric.
For rare-event tasks under class imbalance, p-values alone are insufficient:
they must be paired with AP or F1 to avoid actively misleading conclusions.

\end{multicols}

\begin{multicols}{2}

\paragraph{Summary.}
Three structural findings emerge from the daily equity panel.
\emph{Classification is immune to augmentation}: no generator improves directional
prediction on average: the near-zero baseline leaves no room for variance reduction
to dominate distributional bias.
\emph{Volatility forecasting responds strongly}: all generators are uniformly
significant and $\hat\delta$ grows monotonically with ratio,
with copula families providing the best $R^2$ gains for linear models.
\emph{Targeted augmentation is necessary but not sufficient for rare events}:
targeted deep generative models (DDPM flat, TimeGAN) with tree-based classifiers
(XGB) achieve meaningful AP gains, but the permutation framework, designed for
unconditional augmentation, systematically penalises the distributional shift
that makes targeting effective.
These results underscore that evaluation metrics must be chosen
to match the economic objective of the task.

\section{Empirical Results: Tick Tape}
\vspace{0.2cm}

\noindent
We now turn to the high-frequency tick-by-tick options trade tape. This environment is structurally distinct from the daily equity panel: although the dataset is extremely large in row count, the prediction target is dominated by near-zero returns, exhibits extreme leptokurtosis, and displays severe regime imbalance, with stressed volatility states representing a negligible fraction of observations. Directional prediction at the tick horizon is therefore intrinsically low signal-to-noise, making it a stringent test of whether synthetic augmentation provides incremental predictive information.
\vspace{0.1cm}

\noindent
The number of synthetic samples is set to $m_{\text{extra}} = \mathrm{round}(m_{\text{ratio}} \times n_{\text{train}})$, with augmentation ratios of 0.25 and 0.5. The combined real-plus-synthetic dataset is used to train the “synth” model, while the real-only dataset is used to train the “base” model.
\vspace{0.1cm}

\noindent
To separate the effect of informative synthetic structure from the mechanical effect of increasing sample size, we construct a null augmentation of identical size. The null dataset appends perturbed observations to the real training set, for example by permuting labels or shuffling features, thereby preserving marginal structure while destroying predictive alignment. The model trained on real-plus-null data serves as the null benchmark.
\vspace{0.1cm}

\noindent
For each split, training size, augmentation ratio, and generator, we evaluate three classifier families: logistic regression (L1 and L2 penalties across multiple regularization strengths), Ridge classifier (multiple $\alpha$ values), and random forest (varying depth and minimum leaf size, with a fixed number of trees). All evaluation is strictly out-of-sample on the held-out test window. Predictive performance is measured using log loss, computed per test observation and averaged to obtain $\text{loss}_{\text{base}}$, $\text{loss}_{\text{synth}}$, and $\text{loss}_{\text{null}}$.
\vspace{0.1cm}

\noindent
Statistical significance is assessed using the paired block permutation test described in Section \ref{sec:framework}. Under the null hypothesis that the corresponding risk differential is zero, blockwise sign-flipping preserves temporal dependence while generating an exact finite-sample reference distribution. For each configuration we compute $p_{\text{base}}$ and $p_{\text{net}}$. Low $p$-values (e.g., $<0.05$) indicate statistically significant improvement. When appropriate, Benjamini–Hochberg false discovery rate correction is applied across multiple tests.
\vspace{0.1cm}

\begin{figure*}[!t]
\centering
\includegraphics[width=\linewidth]
{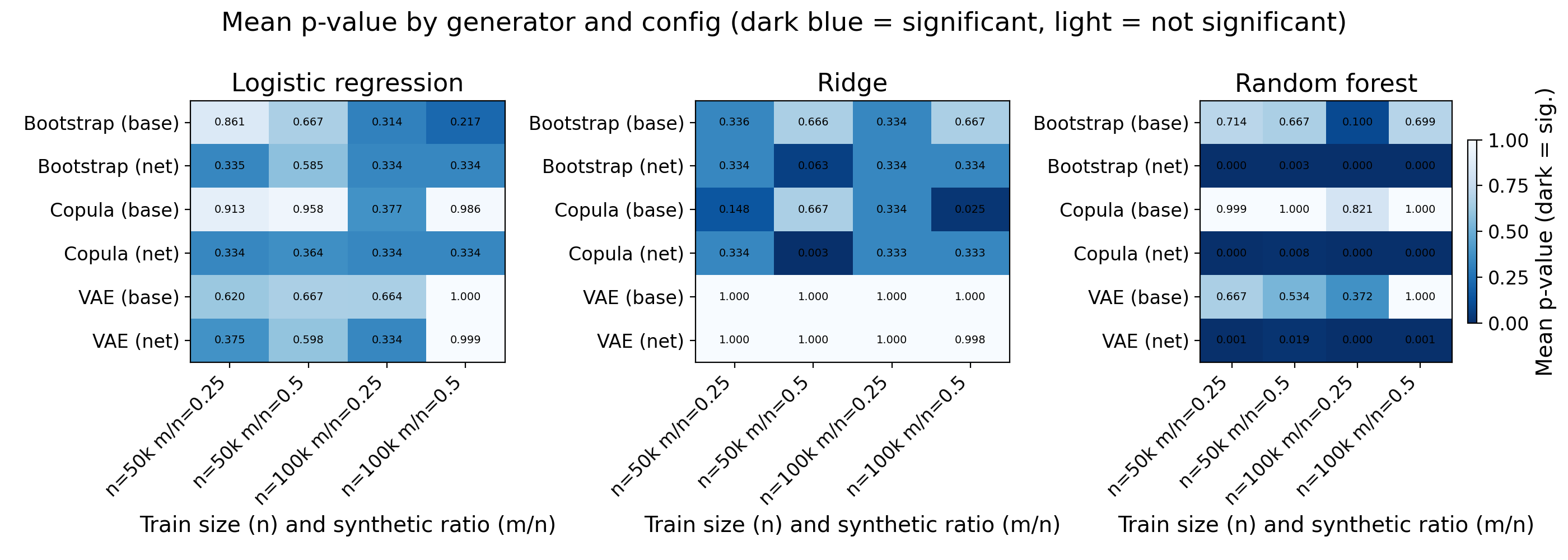}
\caption{Mean block-permutation $p$-values for the tick-by-tick option direction prediction task. Rows correspond to synthetic generators (Bootstrap, Gaussian Copula, VAE) and comparison type (base: synth vs real; net: synth vs null). Columns correspond to training size and augmentation ratio configurations. Panels show Logistic Regression (left), Ridge (center), and Random Forest (right). Darker cells indicate lower mean $p$-values and stronger statistical evidence that synthetic augmentation reduces out-of-sample log loss.}
\label{fig:tick_heatmap}
\end{figure*}

\noindent
Results are aggregated across splits and configurations by computing the mean $p$-value for each generator, comparison type (base or net), and classifier family. Figure~\ref{fig:tick_heatmap} summarizes these results in a single heatmap, where darker cells correspond to lower mean $p$-values and therefore stronger statistical evidence of improvement.
\vspace{0.1cm}

\noindent
The left panel of Figure~\ref{fig:tick_heatmap} shows that logistic regression does not benefit materially from synthetic augmentation. Across generators and augmentation ratios, both $p_{\text{base}}$ and $p_{\text{net}}$ remain well above conventional significance thresholds. The absence of dark cells indicates that low-capacity linear classifiers are unable to extract incremental predictive information from the synthetic distribution in this high-noise setting.
\vspace{0.1cm}

\noindent
The middle panel presents analogous results for the Ridge classifier. With the exception of isolated configurations under the Gaussian copula at smaller training sizes, mean $p$-values remain high and statistical rejection is rare. Overall, regularized linear models exhibit limited sensitivity to synthetic augmentation on the tick tape.
\vspace{0.1cm}

\noindent
In contrast, the right panel reveals a markedly different pattern for the random forest. While synth-versus-real comparisons ($p_{\text{base}}$) are generally not significant, synth-versus-null comparisons ($p_{\text{net}}$) are consistently highly significant across generators and augmentation ratios. The dominance of dark cells for $p_{\text{net}}$ indicates robust rejection of the null hypothesis that synthetic augmentation provides no incremental benefit relative to size-matched perturbations. This pattern holds for Bootstrap, Copula, and VAE generators.
\vspace{0.1cm}

\noindent
The divergence between linear models and random forests suggests that the usefulness of synthetic augmentation in the tick-tape environment is strongly model-capacity dependent. The high-frequency setting features nonlinear interactions among microstructure variables, volatility clustering, and order-flow dynamics. Tree-based ensembles, with their ability to capture local heterogeneity and higher-order interactions, appear capable of leveraging the additional diversity introduced by structured synthetic sampling. In contrast, linear decision boundaries, despite regularization, cannot translate this additional variation into improved generalization.
\vspace{0.1cm}

\noindent
Taken together, the tick-tape results indicate that synthetic augmentation does not uniformly improve predictive performance. Its effectiveness depends critically on the interaction between generator structure and model capacity. In this regime-imbalanced, heavy-tailed environment, gains are detectable primarily when synthetic structure is evaluated relative to a null control and when the predictive model possesses sufficient flexibility to exploit nonlinear features of the augmented distribution.

\section{Conclusion}

This paper develops a structural framework for evaluating synthetic augmentation in financial machine learning by comparing synthetic training sets to a size-matched null augmentation and testing the resulting out-of-sample risk differential with a finite-sample permutation procedure. Across all experiments, the evidence shows that synthetic data is not generically beneficial: its value depends on task predictability, model capacity, regime rarity, and the fidelity of the generator.
\vspace{0.1cm}

\noindent
The daily equity results illustrate this heterogeneity sharply. For directional prediction, augmentation is essentially ineffective: the best risk differential is only $+0.0020$ (Copula Student-$t$ + LogReg), the maximum rejection rate is just $27\%$, and all generator averages are non-positive overall. By contrast, volatility forecasting benefits strongly and uniformly: all $33$ generator--model pairs are positive with $p<0.001$, the best null-relative gain reaches $\hat\delta=+0.137$, and the economically relevant gain versus the unaugmented baseline is about $\Delta R^2=+0.011$ for copula-based augmentation with linear and ridge models (from a baseline $R^2=0.638$). For random forests, baseline performance is already high ($R^2=0.721$), and augmentation adds little in direct $R^2$ terms despite remaining statistically significant. In the rare-event volatility-spike task, spikes represent only $5\%$ of observations (threshold $z \approx 1.925$). There, targeted augmentation improves the domain metric most for DDPM (flat) + XGBoost ($+0.125$ AP) and TimeGAN + XGBoost ($+0.108$ AP), even though balanced logistic regression delivers the lowest permutation p-values with little or no AP improvement; the largest targeted-versus-vanilla p-value gap reaches $\Delta p=-0.964$, showing that unconditional permutation significance can penalize the very distributional shift that makes targeted oversampling useful.
\vspace{0.1cm}

\noindent
The high-frequency tick-tape results reinforce the same message. Synthetic augmentation does not help low-capacity linear models: logistic regression and most ridge configurations remain non-significant, with mean p-values typically between about $0.15$ and $1.00$. In contrast, random forests show strong incremental gains relative to the null benchmark, with net p-values between $0.000$ and $0.019$ across bootstrap, copula, and VAE generators, while the corresponding synth-versus-real comparisons are mostly not significant. Overall, synthetic data should therefore be viewed not as a universal remedy for limited data, but as a conditional tool: it is most useful when the underlying task contains persistent structure that can be smoothed or densified by the generator and when the predictive model is flexible enough to exploit it. In low-signal directional tasks, augmentation is largely neutral or harmful; in rare-event settings, evaluation must be aligned with the economic objective rather than relying on unconditional significance alone.
\vspace{0.1cm}

\noindent
Synthetic data is not a shortcut to predictive power; it is a stress test of whether the signal was there to begin with.

\end{multicols}
\newpage
\bibliographystyle{plainnat}
\bibliography{references}
\end{document}